\documentclass[11pt]{article}
\pdfoutput=1 %
\usepackage{prior}

\usepackage{amsmath,amsfonts,bm}

\def\eqref#1{equation~\ref{#1}}

\def\1{\bm{1}}

\DeclareMathAlphabet{\mathsfit}{\encodingdefault}{\sfdefault}{m}{sl}
\SetMathAlphabet{\mathsfit}{bold}{\encodingdefault}{\sfdefault}{bx}{n}

\tcbuselibrary{most}   %

\newtcolorbox{rough}[1]{
  colback=yellow!15,      %
  colframe=red!70!black,  %
  coltitle=white,
  title=\textbf{#1},
  fonttitle=\bfseries,
  boxrule=1pt,
  arc=3pt,
  breakable
}

\newtcolorbox{figplan}[1][]{colback=gray!4, colframe=gray!55, boxrule=0.8pt, arc=2pt,
  halign=center, valign=center, fonttitle=\bfseries, #1}

\usepackage{etoolbox}
\makeatletter
\patchcmd{\prior@firstfoot}{\textsuperscript{\dag}Corresponding author. Email:}{Corresponding authors:}{}%
  {\PackageWarning{paper}{could not pluralise the corresponding-author label}}
\makeatother

\title{Pretraining Representation Lottery Determines Fine-tuning Generalization}
\shorttitle{The pretraining representation lottery}
\author{Fuming Yang\textsuperscript{*}, Core Francisco Park\textsuperscript{*$\dagger$}}
\authornote{\textsuperscript{*}Equal contribution. \textsuperscript{$\dagger$}Senior contribution.}
\correspondingauthor{Fuming Yang, \texttt{fuming@priorcomputers.ai}; Core Francisco Park, \texttt{core@priorcomputers.ai}}

\begin{document}

\begin{abstract}
Large scale pretraining (PT) followed by fine-tuning (FT) is the central paradigm of modern deep learning, yet how PT shapes FT generalization remains poorly understood. Here, we use a synthetic setup in which the ground-truth concept geometry of entities is known, serving as a controlled testbed to study how representational geometry affects model behavior and how PT representations determine FT generalization. We first show that models differing only in training seed can closely match in PT loss yet vary substantially in FT generalization. We hypothesize that a representational lottery underlies this phenomenon: the random seed dependent geometry formed during PT determines what FT can reach. We confirm this causally. First, injecting the ground-truth concept geometry into the model yields near-ceiling FT generalization, showing that good representations directly drive generalization. Then, distilling a lucky seed's \emph{learned} representation into an unlucky seed unlocks FT generalization, while distilling the same model's behavior does not. Furthermore, representation distillation can merge entities that never interacted during PT, which FT alone cannot. Finally, we show that regularizing representational drift during FT enables naturalistic learning: only models with good representations learn the task, and they generalize, while models with poor representations fail to fit it. Under plain FT, by contrast, models with poor representations fit the training objective by reshaping their pretrained representations. We reproduce the distillation result in Pythia-160M and Pythia-410M, which were each pretrained multiple times with different random seeds, and show that the transferable representation lives in the deep layers at both scales. Overall, PT acts as a lottery over representations, and the representation a model draws determines how it adapts during FT in a way that behavior alone does not explain.
\end{abstract}

\maketitle
\section{Introduction}
\label{sec:intro}
Pretraining (PT) on large-scale data followed by task-aligned fine-tuning (FT) is the main paradigm of modern deep learning \citep{bommasani2021opportunities}.
It is generally believed that the rich representations built during PT are the main
driver of FT performance.
Some recent work has studied the interplay between PT and FT, both in controlled
synthetic settings
\citep{jain2023mechanistically,allen2023physics,zhang2025interplaypretrainingmidtrainingrl,park2026convergent}
and in large language models (LLMs)
\citep{prakash2024finetuningenhancesexistingmechanisms,ward2025reasoningfinetuningrepurposeslatentrepresentations,gekhman2024doesfinetuningllmsnew},
where FT can have exotic effects
\citep{cloud2025subliminallearninglanguagemodels}.
However, our understanding of how PT shapes up a model for FT remains limited,
especially at the representation level.

PT runs that differ only in their random seed offer a controlled way to address this
question. Such runs share the data, the architecture, and the training method, and typically reach
similar PT loss and similar downstream performance
\citep{vanderwal2025polypythiasstabilityoutliersfifty}. However, runs that differ only in their
seed can behave very differently out of distribution even when their in-distribution performance
is nearly identical \citep{jordan2024varianceneuralnetworktraining}. How this seed variance
propagates through FT, whether it decays, persists, or grows, is much less understood: the few
studies that fine-tune models with different PT seeds find that the PT seed still matters after
FT \citep{damour2020underspecificationpresentschallengescredibility,sellam2022multibertsbertreproductionsrobustness},
but do not isolate the cause. To understand why runs with similar PT loss can diverge after FT, we turn to the
underlying premise of the paradigm: the \emph{representations} formed during PT.

\begin{figure}[t]
\centering
\includegraphics[width=\textwidth]{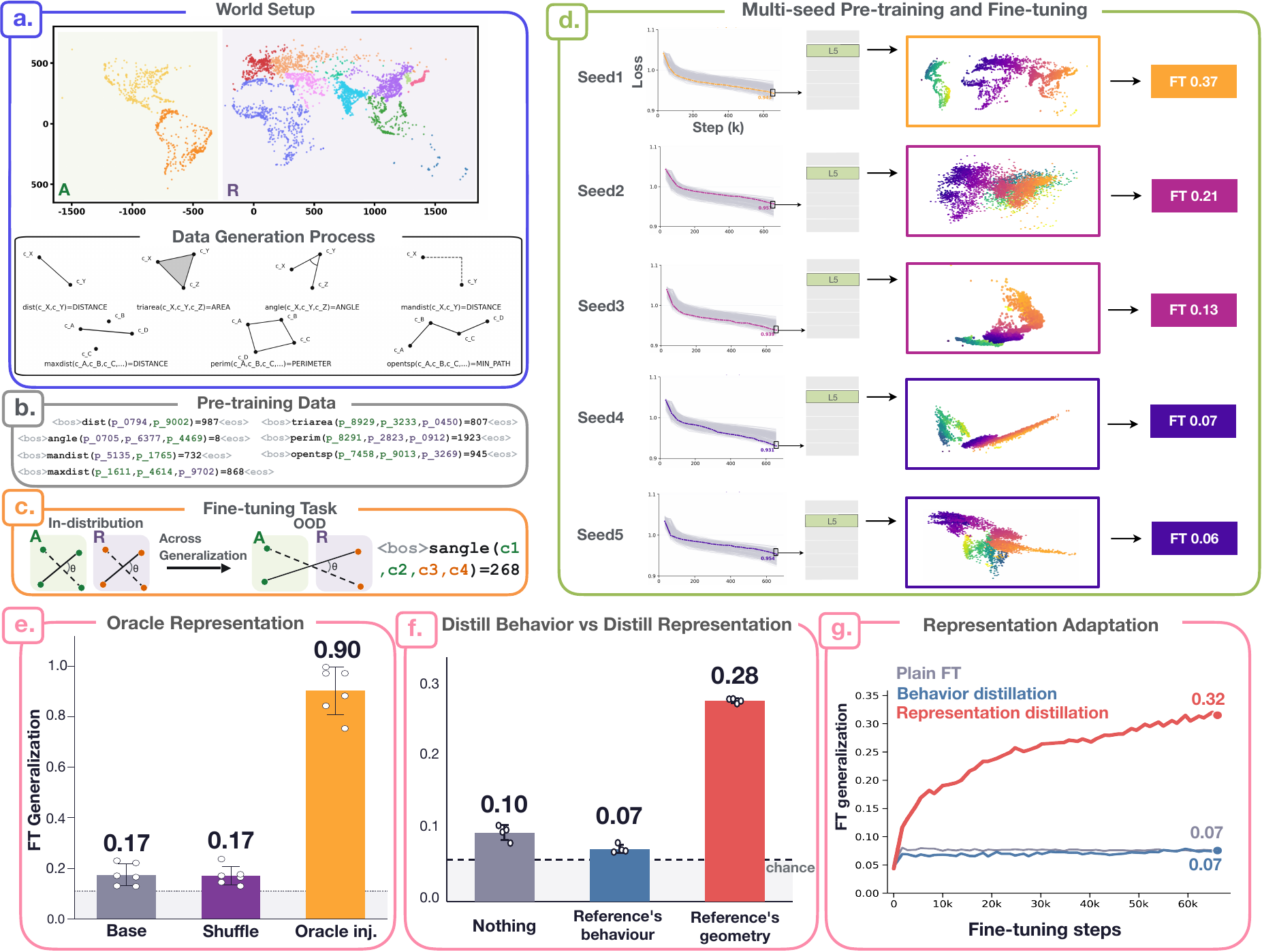}
\caption{\textbf{Overview.} A representation lottery during pretraining determines fine-tuning generalization.
\textbf{a.}~World Setup. $5{,}175$ cities with known geometry constitute the underlying world, and geometric tasks are defined on top of them. The cities are split into two groups, the Americas (A) and the rest of the world (R) (Sec.~\ref{sec:setup}).
\textbf{b.}~Training Data. The pretraining data consist of strings generated by plugging cities into geometric tasks, tokenized at the character level (Sec.~\ref{sec:setup}).
\textbf{c.}~Fine-tuning Task. An unseen fine-tuning task, \texttt{sangle}, is trained on within-group
instances and tested across groups (Sec.~\ref{sec:setup}).
\textbf{d.}~Multi-seed experiments. 
We pretrain multiple runs that differ only in their seed. Different runs form different representations, and their fine-tuning performance varies substantially
(Sec.~\ref{sec:lottery}).
\textbf{e.}~Oracle Representations. Injecting the true coordinates as representations greatly improves FT generalization (Sec.~\ref{sec:oracle}).
\textbf{f.}~Representation Distillation. Distilling a good run's representation into a poor
run improves its FT generalization, while distilling behavior does not help (Sec.~\ref{sec:distill}).
\textbf{g.}~Representation Adaptation. A representation left split during PT can be merged via representation distillation, but not via FT (Sec.~\ref{sec:ortho}).}
\label{fig:overview}
\end{figure}
\textbf{This work.} Investigating this question in real language models is hard.
Pretraining them many times is expensive and, more fundamentally, their representations are
difficult to extract and study beyond comparisons across models such as CKA
\citep{kornblithcka}. We therefore study it in the small-scale, controlled synthetic setup of
\citet{park2026convergent}, in which the ground-truth geometry of the entities is known, the data
are generated procedurally, and the representation of every entity can be extracted efficiently
(Sec.~\ref{sec:setup}).
All models we study are Transformers trained on token sequences: our synthetic setup and Pythia-160M and Pythia-410M. Other architectures and modalities, such as CNNs on images, are not tested.

The setup reproduces the phenomenon in question: runs with the same PT loss fine-tune very
differently. Rather than correlating properties of the pretrained representation with downstream outcomes
\citep{garrido2023rankmeassessingdownstreamperformance}, we intervene on the representation directly,
with controls that rule out the alternative explanations. These experiments show that the
representation a model forms during PT strongly determines what FT can achieve, and that this
representation can be distilled into another model to lift its FT generalization, while
distilling the model's behavior does not. Regularizing the representation during FT then lets a
model learn a task only if its representation already supports it. The distillation result reproduces in Pythia-160M and Pythia-410M. \textbf{Overall, our
contributions are:}
\begin{enumerate}[leftmargin=*, topsep=2pt, itemsep=1pt, parsep=0pt, partopsep=0pt]
    \item \textbf{Pretraining seeds with similar loss fine-tune very differently
    (Sec.~\ref{sec:lottery}, Fig.~\ref{fig:lottery})}: 
    Across $278$ models trained with different PT seeds, FT generalization spans a factor of seven for similar PT loss. Furthermore, the FT seed has a much smaller effect than the PT seed.
    \item \textbf{The representation formed during pretraining strongly determines how fine-tuning generalizes, and can be distilled into other models (Sec.~\ref{sec:causal},
    Fig.~\ref{fig:oracle}, \ref{fig:distill}, \ref{fig:ortho})}: We show that oracle representations support FT at near
    ceiling performance; a good seed run's learned representation can be distilled into a poor one and transmit its
    fine-tunability, while distilling behavior does not; and a joint representation can be installed in a
    model whose PT made it impossible, which FT cannot do.
    \item \textbf{Representation regularization reveals the true learnability axis
    (Sec.~\ref{sec:hidden}, Fig.~\ref{fig:hidden})}: When the representation is regularized during FT, only models whose representation supports the task learn it, allowing ``naturalistic learning''.
    \item \textbf{Confirmation of representation distillation in real language models (Sec.~\ref{sec:llm},
    Fig.~\ref{fig:llm})}: We confirm representation distillation in Pythia-160M and Pythia-410M. Distilling a strong PT seed's
    representation into a weak one lifts its two-hop OOD generalization above a self-distillation
    control ($0.256 \to 0.403$ at 160M; at 410M in $22$ of $24$ deep-layer settings, e.g.\ $0.417 \to 0.528$ for one recipient at layer 22), but only from deep
    layers; early layers transfer nothing.
\end{enumerate}
In this work, we hope to deliver \textit{some} insights into this question in \textit{at least a small synthetic domain}.
We discuss related work in App.~\ref{sec:related}.

\section{Setup}
\label{sec:setup}
Fig.~\ref{fig:overview} illustrates our experimental setup.

\textbf{World setup.} We build on the synthetic setup introduced in \citet{park2026convergent}. Here, the underlying world consists of $5{,}175$ cities, whose flattened coordinates are the two
latent coordinates $(x, y)$ defining the entity. Cities are split into the Americas (A) and the rest of the world
(R) (Fig.~\ref{fig:overview}a.) in order to define OOD generalization for FT. The PT data are relational statements produced by seven geometric tasks over the cities, for example \texttt{dist(p\_6400,p\_8995)=728}
(Fig.~\ref{fig:overview}a, b; App.~\ref{app:training}). A city appears only as a character string, never with its coordinates, so the model must build its representation from the relational data it is involved in. Statements mix cities from both groups freely, except in the \emph{ortho} control of Sec.~\ref{sec:ortho}, where no statement relates A to R. We pretrain a 6-layer Transformer (Qwen2 architecture \citep{yang2024qwen2technicalreport}, $d=128$) for $656$k steps, repeated over $278$ random seeds randomizing only the model initialization and data order (Fig.~\ref{fig:overview}d, App.~\ref{app:grid}).

\textbf{Fine-tuning task.} The FT task, which is never shown during PT, is \texttt{sangle}$(c_1,c_2,c_3,c_4)$ which should output the signed angle between the segments $c_1{\to}c_2$ and $c_3{\to}c_4$. We evaluate the accuracy as correctness within $10^{\circ}$ (Fig.~\ref{fig:overview}c). The FT data contain only within-group instances, whose
four cities all come from A or all from R. In-distribution (ID) accuracy is measured on held-out
within-group instances, and \emph{FT generalization} is the accuracy on cross-group instances,
which combine cities from A and R (App.~\ref{app:comps}).

\textbf{Representation readout.} We read out a model's representation of an entity (city) as the residual representation after the 5\textsuperscript{th} Transformer layer (App.~\ref{app:probe}). These representations are shown in Fig.~\ref{fig:overview}d. We later use these representations to distill or regularize FT.

\section{The pretraining seed is a lottery for fine-tuning}
\label{sec:lottery}
\begin{figure}[t]
\centering
\includegraphics[width=0.82\textwidth]
{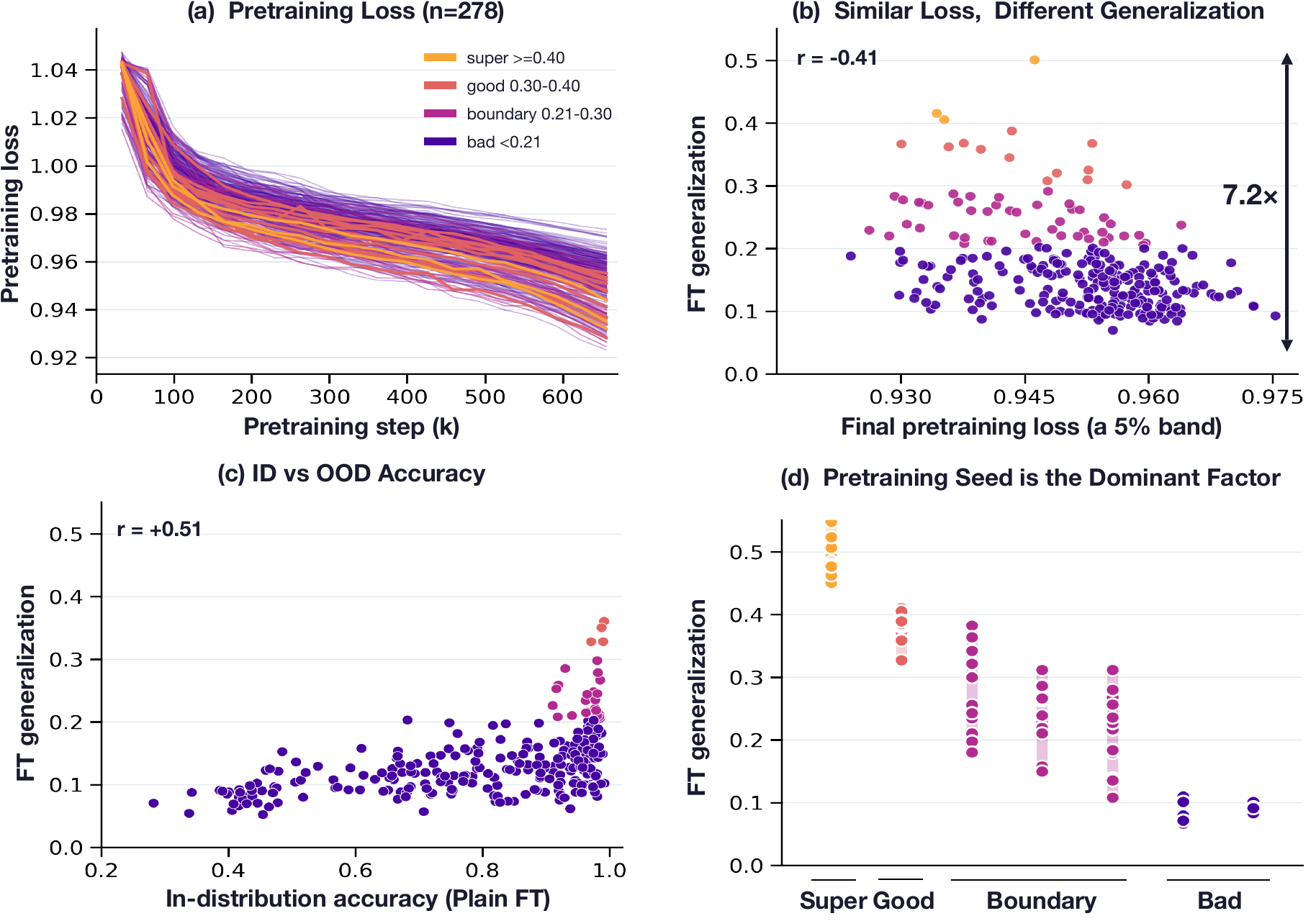}
\caption{\textbf{Same pretraining loss, different fine-tuning.} Runs that differ only in their PT
seed reach similar PT loss but fine-tune very differently.
\textbf{a.}~Pretraining Loss. PT loss curves of all $278$ runs, colored by tier; the tiers do not cleanly separate by PT loss (App.~\ref{app:curves}).
\textbf{b.}~Similar Loss, Different Generalization. Within a $5\%$ band of final PT loss, FT
generalization varies $7.2$-fold and correlates only weakly with the loss ($r=-0.41$).
\textbf{c.}~ID vs OOD Accuracy. After plain FT, most runs fit the FT task in distribution,
including runs that do not generalize OOD.
\textbf{d.}~Pretraining Seed is the Dominant Factor. Seven PT runs, each fine-tuned with
fourteen FT seeds: \emph{super}, \emph{good}, and \emph{bad} runs keep their tier, and only \emph{boundary} runs show significant spread
(App.~\ref{app:ftseed}).}
\label{fig:lottery}
\end{figure}

We pretrain $278$ models that differ only in their initialization and data-order seeds
(Fig.~\ref{fig:lottery}a), then fine-tune each of them on the \texttt{sangle} task and measure their FT
generalization accuracy. Although the final PT losses of all runs fall within a $5\%$ band, their FT
generalization (the peak reached during FT; App.~\ref{app:notation}) ranges from $0.07$ to $0.50$, a factor of seven (Fig.~\ref{fig:lottery}b). This
variation persists when we control for the PT loss as precisely as possible (Fig.~\ref{fig:isoloss}). For ease of discussion, we group the
runs into four tiers by FT generalization: \emph{super} ($\geq 0.40$, $3$ runs), \emph{good}
($0.30$--$0.40$, $12$ runs), \emph{boundary} ($0.205$--$0.30$, $47$ runs), and \emph{bad} (within
the null band, $216$ runs).

While the PT loss retains a weak correlation with FT
generalization ($r=-0.41$, $R^{2}=0.17$), it cannot select the runs that matter: of the $27$
runs in the lowest-loss decile, only $2$ are \emph{super} or \emph{good}, indistinguishable from a random decile at the $5.4\%$ base rate ($15$ of $278$; App.~\ref{app:stats}). Additionally, we find that the in-distribution FT accuracy is also a poor predictor of generalization: the median run reaches $0.89$, and runs that never generalize fit the FT task in-distribution as well as runs that do generalize (Fig.~\ref{fig:lottery}c); a result we will revisit in Sec.~\ref{sec:hidden}.

For completeness, we also fine-tune seven PT runs with fourteen FT data-order seeds each
(Fig.~\ref{fig:lottery}d; App.~\ref{app:ftseed}). For \emph{super}, \emph{good}, and \emph{bad} runs, the spread
caused by the FT seed is much smaller than the spread across PT seeds and only \emph{boundary} runs are
sensitive to the FT seed. Furthermore, we find that a good run is a joint property of its initialization
\emph{and} its data order: crossing ten runs' initializations with their data orders, a run's original
score predicts neither how its initialization does with other data orders nor how its data order does
with other initializations (App.~\ref{app:grid}).

\textbf{Overall, how well a pretrained model fine-tunes is set by its PT seed, in a way that
neither the PT loss nor in-distribution accuracy reveals: pretraining is a lottery for
fine-tuning.}

\section{The pretraining lottery is over representations}
\label{sec:causal}
\label{sec:reps}

The results of Sec.~\ref{sec:lottery} show that neither the PT loss nor in-distribution
accuracy explains FT generalization. We therefore turn to the representation itself and
hypothesize:

\textbf{Hypothesis.} \emph{Pretraining is a lottery over representations: the representation a
run draws determines its FT generalization, and it can be transferred between models, while
behavior cannot.}

A linear probe of the pretrained model, taken from a single forward pass before any FT, gives some
further support: its top decile holds $7$ of the $15$ good runs, well above the $5.4\%$ base rate,
though far from reliably (App.~\ref{app:signature}, \ref{app:stats}).

This evidence is correlational. We therefore intervene on the representation directly and show
that (i) the ground-truth representation is sufficient for near-ceiling FT generalization
(Sec.~\ref{sec:oracle}); (ii) a good run's representation can be distilled into a poorly
generalizing run and elicits generalization, while distilling its behavior does not
(Sec.~\ref{sec:distill}); and (iii) a joint representation can be distilled into a model whose
PT kept the two entity groups apart, which, again, FT cannot achieve (Sec.~\ref{sec:ortho}).

\subsection{Oracle representations fine-tune at ceiling generalization}
\label{sec:oracle}
\begin{figure}[t]
\centering
\includegraphics[width=\textwidth]{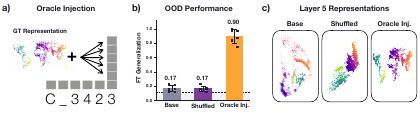}
\caption{\textbf{Oracle representations fine-tune at ceiling.} Supplying the ground-truth representation
lifts FT generalization to near ceiling; supplying one with the wrong geometry does not.
\textbf{a.}~Ground-truth representation input. Each city's true coordinates are projected into
the residual stream and added at the city's last id token, at every layer after the first
(App.~\ref{app:oracle}).
\textbf{b.}~OOD Performance. FT generalization without injection (Base), with shuffled
coordinates (Shuffle), and with the true coordinates (Oracle Inj.); bars are means over PT seeds
with standard errors, dots individual PT seeds, and the dotted line chance.
\textbf{c.}~Layer-5 Representation. Layer-5 representations of the three models, colored by true
longitude.}
\label{fig:oracle}
\end{figure}

First, we ask whether a good representation is \emph{sufficient} for FT generalization. We give the
model the ground-truth (GT) representation directly: a learned linear map transforms each city's true
coordinates $(x,y)\in\mathbb{R}^2$ into a $128$-dimensional vector, which we add to the residual
stream at the city's last id token, at the input of every layer after the first, throughout PT and
FT (Fig.~\ref{fig:oracle}a; App.~\ref{app:oracle}). With the GT representation given, FT
generalization rises to near ceiling, from $0.17$ to $0.90$
(Fig.~\ref{fig:oracle}b). This experiment uses a separately implemented version of the setup (App.~\ref{app:oracle}), so
only the comparison between its three conditions is meaningful.

Injected coordinates both mark each city as a single entity and carry the world's geometry. To disambiguate the gains, we also inject shuffled coordinates where each city gets a consistent vector,
but the geometry is wrong. This control fits the FT task in distribution yet stays at the base level
($0.17$; Fig.~\ref{fig:oracle}b), and thus we confirm the gains originate from having the correct geometry. The layer-5 representations show the supplied map in place (Fig.~\ref{fig:oracle}c); the
oracle model reads it from the injection instead of building its own (App.~\ref{app:oracle}), demonstrating that the correct representation is indeed in use.

\textbf{Overall, a representation with the correct geometry is sufficient to make FT generalize well.}

\subsection{Distilling a good representation allows fine-tuning generalization}
\label{sec:distill}
\begin{figure}[t]
\centering
\includegraphics[width=0.85\textwidth]{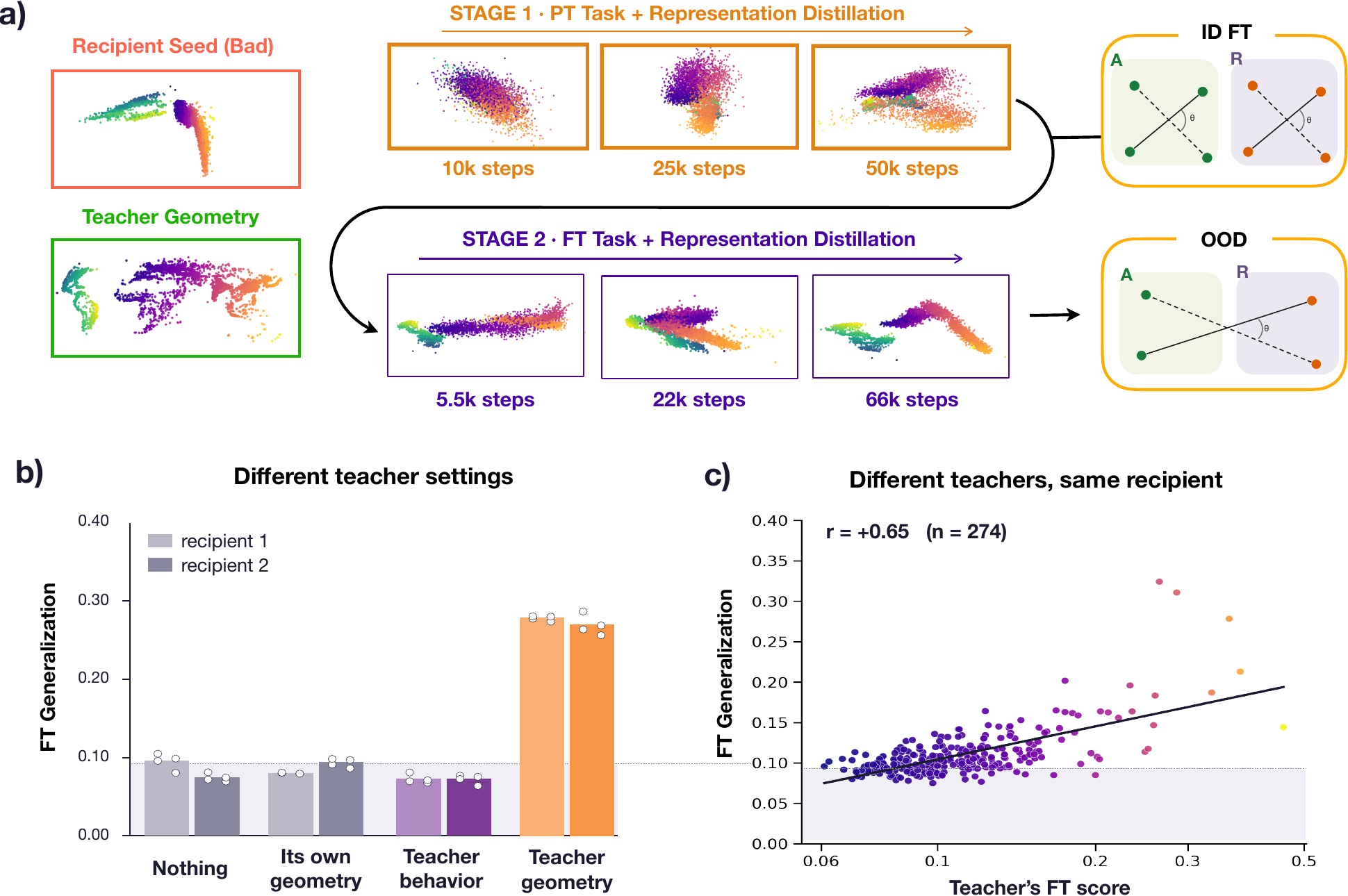}
\caption{\textbf{Distilling a good representation transfers FT generalization.} A good run's layer-5
representation can be distilled into a poor run to improve its FT generalization while distilling the same teacher's behavior does not.
\textbf{a.}~Representation distillation. Stage 1: continued pretraining on ordinary PT text but with the loss pulling the recipient's layer-5 activations toward the teacher's.
Stage 2: FT with the distillation loss still active, to persist the installed geometry through FT.
\textbf{b.}~FT Generalization per Recipient. Two bad recipients' FT generalization after plain FT (Nothing),
distillation toward their own pretrained states, distillation of the teacher's behavior, and distillation of the
teacher's representation; dots are FT seeds.
\textbf{c.}~274 Teachers into One Recipient. Distilling each of $274$ runs into the same
bad-tier recipient, the recipient's FT generalization tracks the teacher's regularized FT generalization
($r=0.65$ against the logarithm of the teacher's score, as plotted; $0.70$ on a linear scale).}
\label{fig:distill}
\end{figure}

Having established that a good representation is sufficient, we now show that it is also
transferable: a good representation, the winning ticket of the PT lottery, can be moved into a run
that drew a poor one. We distill the representation of a \emph{super} run, the teacher, into a
bad run, the recipient (Fig.~\ref{fig:distill}a). The procedure resembles representation
distillation and alignment methods from the literature
\citep{romero2015fitnets,tian2022contrastiverepresentationdistillation,yu2025representationalignmentgenerationtraining},
but what we transfer is the recipient's later FT generalization rather than its current
behavior. In a first stage, on ordinary PT text $x$ with no task data, a representation
distillation (RD) loss attracts the recipient's layer-5 states toward the teacher's,
\[
\mathcal{L}_{\mathrm{RD}}(\theta) \;=\;
\big\lVert h_{5}(x;\theta) - h_{5}(x;\theta_{\mathrm{teacher}}) \big\rVert^{2},
\]
which installs the teacher's representation in the recipient. In a second stage, the recipient is
fine-tuned with the RD loss kept on.\footnote{Both stages are necessary; see the ablation in
App.~\ref{app:routes}.} We compare RD with distilling the teacher's behavior by standard
distillation \citep{hinton2015distilling}, which matches the teacher's next-token distribution on
the same PT corpus.

RD lifts both recipients from the floor to $0.28$ and $0.27$, while plain FT and distillation
toward the recipient's own pretrained representation stay in the null band
(Fig.~\ref{fig:distill}b). Distilling the teacher's behavior instead reduces the KL divergence to
the teacher ($0.06 \to 0.03$ and $0.10 \to 0.02$) and raises top-$1$ next-token agreement from
$0.81$ to $0.83$--$0.84$, yet FT generalization stays in the null band ($0.07$;
Fig.~\ref{fig:distill}b; App.~\ref{app:behavdist}). The two models already agree on most of the
corpus and agree more after behavior distillation; what separates them is the representation from
which those outputs are read, and only copying that representation transfers the ability. This is in
line with subliminal learning, where training on a teacher's outputs transmits its traits only
when teacher and student share a base model \citep{cloud2025subliminallearninglanguagemodels}.

To verify that RD transfers the teacher's own generalization, we distill each of $274$ runs as
teacher into the same recipient. The recipient's FT generalization tracks the teacher's regularized FT generalization ($r=0.70$, and
$0.65$ against the logarithm of the teacher's score as plotted in Fig.~\ref{fig:distill}c), and the best teachers lift the recipient well out of the null band
(App.~\ref{app:transplant}).

\textbf{Overall, a good representation can be distilled from one model into another and brings
its FT generalization along, while distilling the same model's behavior does not.}

\subsection{Distillation can join representations that fine-tuning cannot}
\label{sec:ortho}

\begin{figure}[t]
\centering
\includegraphics[width=\textwidth]{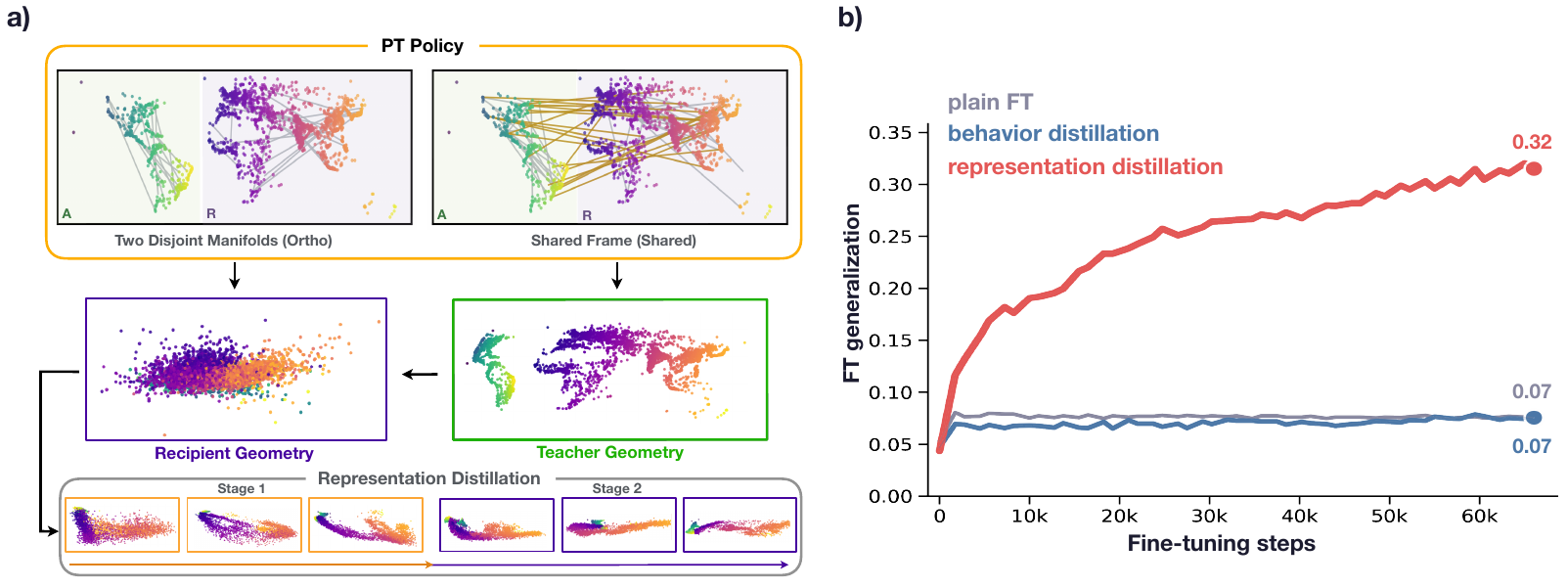}
\caption{\textbf{A joint representation can be installed where pretraining made it impossible.}
\textbf{a.}~An \emph{ortho} model is trained on data where cities in A never interact with cities in R, and thus doesn't have a common representation space relating both groups. Stage 1 pulls its layer-5 representations toward those of a teacher pretrained with mixed groups and a joint map assembles in the recipient; Stage 2 applies the RD loss through FT as in Sec.~\ref{sec:distill}
\textbf{b.}~FT generalization of the \emph{ortho} recipient over FT: it rises only with representation
distillation, and stays at the floor with behavior distillation or plain FT.}
\label{fig:ortho}
\label{fig:causal}
\end{figure}

Finally, we show that RD enables a stronger form of generalization: relating two groups of
cities that never interacted during PT. For this experiment, we pretrain \emph{ortho} models, in which no PT data row relates a city in A to a city in R. These models learn a separate map for each group, and the data
constrain how the two maps are placed relative to each other only weakly (Fig.~\ref{fig:ortho}a;
App.~\ref{app:ident}). No FT run from an \emph{ortho} model reaches the $0.205$ threshold, i.e., all are in the \emph{bad} tier ($0.05$--$0.20$ over $30$ runs; App.~\ref{app:xorseeds}).

Fine-tuning an \emph{ortho} model on all seven \emph{PT tasks} with cross-group data, which supplies exactly the missing relations, but only after PT, smears the two maps into each other without merging them, and FT generalization stays in the null band ($0.11$; $0$ of $25$ runs exceed it;
App.~\ref{sec:xor}). In other words, unrelated representation spaces formed during PT are not easily related during FT.

RD from a teacher pretrained with mixed groups, using the procedure of Sec.~\ref{sec:distill}
(Fig.~\ref{fig:ortho}a), does relate the two groups: FT generalization rises well above chance,
while plain FT and behavior distillation from the same teacher stay near the floor
(Fig.~\ref{fig:ortho}b). The FT data still contain only within-group instances, so the
cross-group generalization can only come from the distilled representation.

\textbf{Overall, we find that a joint representation of entities that never interacted during PT can be
distilled into a model, while fine-tuning on the missing relations cannot build it.}

\section{Representation regularization reveals a learnability axis}
\label{sec:hidden}
\label{sec:dynamics}

\begin{figure}[t]
\centering
\includegraphics[width=\textwidth]{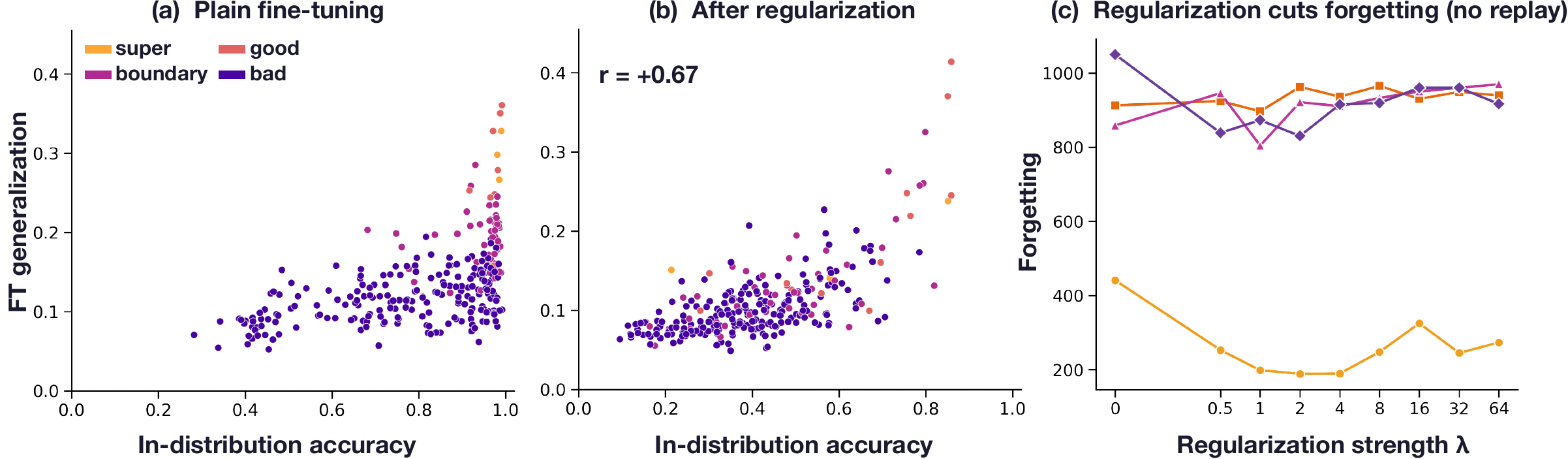}
\caption{\textbf{Representation Regularization reveals learnability.} Under representation regularization, only runs with a good representation learn the task, in distribution and out of distribution, while runs with a bad representation learn less.
\textbf{a.}~Plain FT. In-distribution accuracy against FT generalization for all $278$ runs,
colored by tier: most runs are pressed against the in-distribution ceiling (median $0.89$) whether
or not they generalize ($r=+0.51$).
\textbf{b.}~Regularized FT. The same runs under representation regularization: in-distribution
accuracy spreads over $[0.10, 0.86]$ and tracks generalization ($r=+0.67$).
\textbf{c.}~Forgetting. Forgetting of the PT tasks against the regularization strength $\lambda$:
representation regularization cuts forgetting for the super run; the other three are unchanged, with no replay
data.}
\label{fig:hidden}
\end{figure}

Given that the PT lottery is over good representations, why do models with poor representations still
fit the FT task in distribution (Fig.~\ref{fig:lottery}c)? First, we observe that across the $278$ runs, the total parameter change during plain FT anti-correlates with FT generalization ($r=-0.40$): the runs that generalize are the runs that changed least in parameter space, and the difference is localized to the layers that carry
the representation (App.~\ref{app:reshape} and~\ref{app:layers}). From this observation we hypothesize:

\textbf{Hypothesis.} \emph{Gradient descent tends to fit the FT task by reshaping the pretrained
representation, whether or not the task requires it.\footnote{In our case, this task does not require an entity representation change as no world cities are moving.} What distinguishes a good representation from a bad one is whether the task can be learned without reshaping the representation: under representation regularization, which limits how fast the representation can change during FT, a good
representation can still learn the task, while a poor one
cannot fit the task at all.}

In order to test this, we regularize the representations during FT. \emph{Representation regularization} applies the RD loss of
Sec.~\ref{sec:distill} with the model's own layer-5 states as the target, reset periodically (details in App.~\ref{app:training}; additional experiments in App.~\ref{app:anchorabl}).

Comparing each PT seed's in-distribution accuracy and FT generalization under plain FT
(Fig.~\ref{fig:hidden}a) and under representation regularization (Fig.~\ref{fig:hidden}b), we find
that regularization reveals the true generalization potential of each seed. Under plain FT,
in-distribution accuracy sits near the ceiling for most runs (median $0.89$) and says little about
generalization ($r=+0.51$). Under regularization, runs with bad representations now sit low on
both axes while good runs do well on both: in-distribution accuracy spreads from $0.10$ to $0.86$
and tracks generalization ($r=+0.67$; App.~\ref{app:stats}).

Representation regularization also reduces catastrophic forgetting of the PT tasks without any
replay data, but only partially: forgetting drops by about half for one run and is unchanged for
the others (Fig.~\ref{fig:hidden}c; App.~\ref{app:reshape}).

\textbf{Overall, representation regularization enforces what we call \emph{naturalistic
learning}: learn the task if it can be learned with the representation you have, and refuse a fit
that would destroy it. Learnability is the hidden axis.}

\section{Confirmation in language models}
\label{sec:llm}

\begin{figure}[t]
\centering
\includegraphics[width=0.9\textwidth]{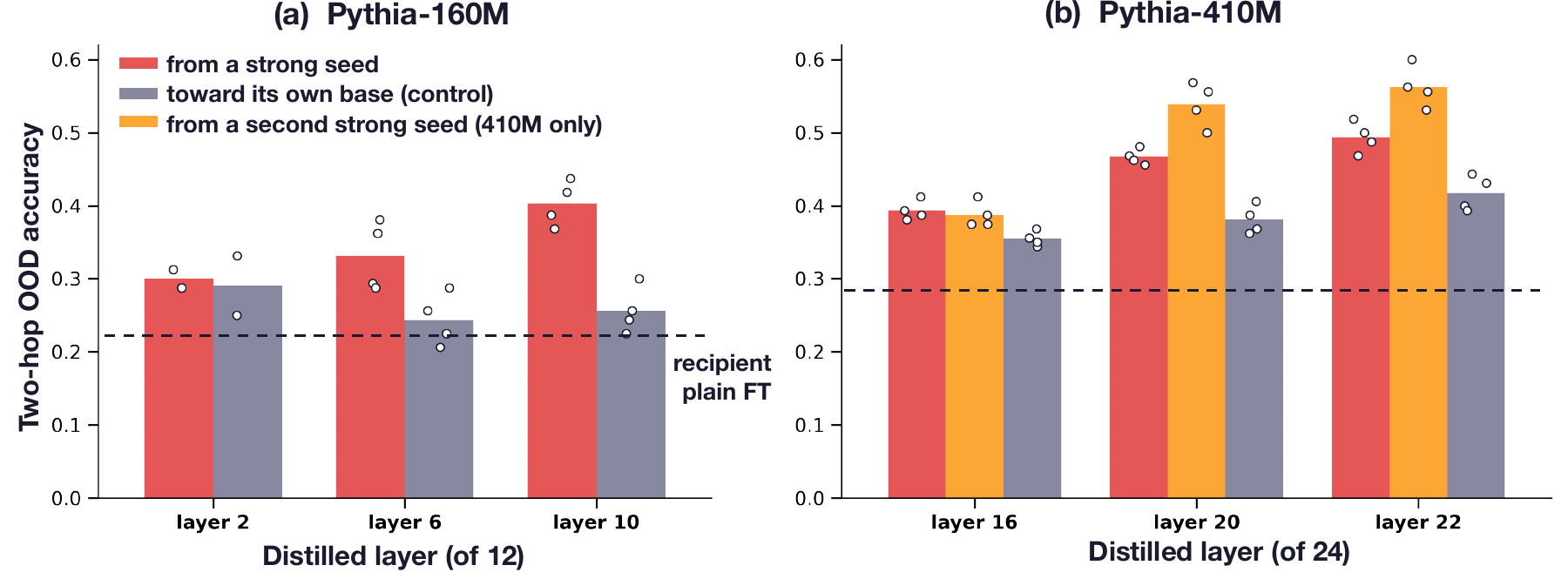}
\caption{\textbf{Representation distillation transfers to Pythia.} Representation distillation
from a strong seed (coloured) beats plain FT (dashed line), but only from deep layers;
representation regularization toward the weak seed's own base states (gray) helps at 160M and
for the first of the two 410M recipients.
\textbf{a.}~Pythia-160M. The worst of ten seeds, distilled at layers 2, 6, and 10 of its 12
layers. Plain FT gives $0.222$; representation regularization gives $0.244$--$0.291$;
representation distillation at layer 10 gives $0.403$, while layer 2 adds nothing beyond
regularization.
\textbf{b.}~Pythia-410M. The worst seed, distilled from two strong seeds at weight $0.1$.
Bars are means and dots individual FT seeds.}
\label{fig:llm}
\end{figure}

We test whether the distillation result holds in language models, at two scales
(App.~\ref{app:llm}).

\textbf{Models and task.} Pythia-160M and Pythia-410M each have ten public PT runs that
differ only in random seed \citep{biderman2023pythiasuiteanalyzinglarge,vanderwal2025polypythiasstabilityoutliersfifty}, the language-model analogue of our
$278$ synthetic runs. We fine-tune on a synthetic two-hop relational world of invented people, cities, and
countries: the corpus states every one-hop fact, and the OOD test asks the two-hop question for
held-out people whose one-hop facts were trained but never composed (corpus, FT protocol, and
seed selection in App.~\ref{app:llm-setup}). The base models answer none of them.
Composing two separately learned facts is a known weakness of language models
\citep{press2023measuringnarrowingcompositionalitygap,allenzhu2024physicslanguagemodels32}.

\textbf{Distillation transfers to language models.} The seeds already differ on the two-hop
OOD test under plain FT ($0.21$--$0.32$ at 160M, $0.30$--$0.54$ at 410M). At 160M we take the
worst seed and apply both interventions of Secs.~\ref{sec:distill} and~\ref{sec:hidden}.
Representation regularization toward the seed's own states raises its OOD accuracy from
$0.222$ under plain FT to $0.244$--$0.291$, at every layer tried. Representation distillation
from a strong seed's layer-10 representation raises it to $0.403$, above regularization alone
at the same layer ($0.256$; four FT seeds per arm; every treated run exceeds every control
run; Fig.~\ref{fig:llm}a; layers, weights, and seed counts in App.~\ref{app:llm-setup}).

\textbf{The same transfer at 410M.} We distill from two strong seeds into two weak ones, at
three depths and three distillation weights, four FT seeds per cell
(Table~\ref{tab:p410}). Regularizing
the recipient toward its own base states, with no reference, exceeds plain FT in $12$ of the
$18$ cells (every cell for the first recipient, and layer 16 for the second). Representation
distillation exceeds plain FT in all $36$ cells and, at layers 20 and 22 of the model's 24,
lifts the first recipient from $0.278$ to $0.47$--$0.56$ and the second from $0.370$ to
$0.48$--$0.55$, at or above the reference seeds' own scores and above the self-regularization
control in $22$ of the $24$ cells with no overlap between treated and control runs. At layer
16, as at layer 2 of the smaller model, the reference adds little beyond regularization alone
(Fig.~\ref{fig:llm}b). At shallow layers, or at distillation weights of $0.5$ and above, the
graft falls below the control, so the transfer needs a deep layer and a light touch (weights).

\section{Discussion}
\label{sec:discussion}
\textbf{Summary.} We showed, in a controlled setup, that FT generalization is mostly decided by the entity representation formed during PT. The quality of the representation, in terms of how it supports FT, seems to be a lottery over the PT seed, and is hard to predict from loss or in-distribution behavior. We found that interventions on the representation can move FT generalization, while distilling behavior does not.

\textbf{Loss of Representational Plasticity.} A broader problem beyond the lottery is that representations are nearly fixed via PT. As we saw in Sec.~\ref{sec:ortho}, a PT run that did not relate some groups of data cannot easily rearrange its representations to support FT generalization, and only a teacher representation could perform the surgery. This connects to a broader problem where representations in neural networks do not change easily after FT \citep{dohare2024lossofplasticity}.

\textbf{Limitations.} Our study has several limitations. First, our findings are mostly synthetic. The main causal results come from a synthetic world, chosen because it provides a controllable ground truth; nevertheless we do reproduce some of the distillation result on Pythia with 160M and 410M parameters on a two-hop reasoning task. Whether
the lottery and the interventions hold in bigger pretraining at scale is unknown. Second, we did not find a single function that predicts, directly from a pretrained representation, how well a run
will fine-tune: geometric summaries barely correlate with the outcome (App.~\ref{app:metrics}),
and the best readout, while predictive, misses most good runs. Our interventions show that the
representation determines FT generalization, but not which property of it does. Third, the
strongest interventions work only partially: distillation lifts poor runs to about $0.28$, far
below the oracle's $0.90$, and regularization must be tuned between allowing destructive reshaping
and a rigid freeze that costs good runs their generalization (App.~\ref{app:anchorabl}).

\clearpage

\bibliography{iclr2027_conference}

\appendix

\noindent Table~\ref{tab:appmap} links the appendix sections to the main text.

\begin{table}[h]
\centering
\caption{\textbf{Appendix guide.}}
\label{tab:appmap}
\footnotesize\begin{tabular}{lll}
\hline
Appendix & Topic & Main-text reference \\
\hline
\ref{sec:related} & Related work & all \\
\hline
\multicolumn{3}{l}{\textit{Experimental setup}} \\
\ref{app:notation} & Notation and terminology & all \\
\ref{app:training} & Synthetic model and training & \S\ref{sec:setup} \\
\ref{app:comps} & FT generalization metric & \S\ref{sec:setup}, \S\ref{sec:hidden} \\
\ref{app:probe} & Representation extraction and probing & \S\ref{sec:setup}, Fig.~\ref{fig:overview} \\
\ref{app:llm} & Language-model setup & \S\ref{sec:llm} \\
\hline
\multicolumn{3}{l}{\textit{Additional experiments}} \\
\ref{app:curves} & Pretraining curves & \S\ref{sec:lottery} \\
\ref{app:ftseed} & Variation across FT seeds & \S\ref{sec:lottery} \\
\ref{app:grid} & Initialization and data order & \S\ref{sec:lottery} \\
\ref{app:signature} & Step-0 answer readout & \S\ref{sec:reps} \\
\ref{app:metrics} & Selection by pretrained-model metrics & \S\ref{sec:reps} \\
\ref{app:oracle} & Oracle injection & \S\ref{sec:oracle} \\
\ref{app:routes} & Representation distillation ablations & \S\ref{sec:distill} \\
\ref{app:transplant} & Distillation across seeds & \S\ref{sec:distill} \\
\ref{sec:xor} & Across-generalization under other PT policies & \S\ref{sec:ortho} \\
\ref{app:xorseeds} & Replications across seeds & App.~\ref{sec:xor} \\
\ref{app:ident} & Cross-group identifiability & App.~\ref{sec:xor} \\
\ref{app:reshape} & Fixed-target regularization and forgetting & \S\ref{sec:hidden} \\
\ref{app:layers} & Per-layer parameter change & \S\ref{sec:hidden} \\
\ref{app:anchorabl} & Fixed-target and periodic regularization & \S\ref{sec:causal}, \S\ref{sec:hidden} \\
\ref{app:stats} & Statistical notes & \S\ref{sec:lottery}--\S\ref{sec:hidden} \\
\ref{app:behavdist} & Behavior distillation & \S\ref{sec:distill} \\
\hline
\multicolumn{3}{l}{\textit{Visualizations}} \\
\ref{app:gallery} & Representation gallery & \S\ref{sec:reps} \\
\hline
\end{tabular}
\end{table}

\section{Related work}
\label{sec:related}

\textbf{Seed variance.} Runs with the same training procedure but different seeds can match on loss and
benchmarks and still generalize differently \citep{juneja2022linear,mccoy2020bertsfeathergeneralizetogether,damour2020underspecificationpresentschallengescredibility}. FT seeds alone can substantially change downstream accuracy, and the
variance has been traced to weight initialization and data order
\citep{dodge2020finetuningpretrainedlanguagemodels} and to optimization instability \citep{mosbach2021stabilityfinetuningbertmisconceptions}.
Suites of models pretrained with different seeds enable study of PT variance
\citep{sellam2022multibertsbertreproductionsrobustness,vanderwal2025polypythiasstabilityoutliersfifty}.
Networks with the same output behavior can have markedly different internal representations
\citep{kumar2025questioningrepresentationaloptimismdeep}.
We identify the pretrained representation underlying this variance and transfer it between runs.
The lottery ticket hypothesis
\citep{frankle2019lottery} attributes training success to a sparse subnetwork present at initialization;
in our experiments, initialization alone does not explain FT outcomes (App.~\ref{app:grid}),
whereas pretrained representations predict them before FT. Seed-to-seed differences have also been
characterized through linear mode connectivity \citep{frankle2020linear}.

\textbf{How pretraining supports fine-tuning.} \citet{neyshabur2020being} ask what transfer
learning transfers, and \citet{kumar2022finetuningdistortpretrainedfeatures} show that
FT can distort pretrained features and degrade out-of-distribution performance.
Models with equal PT loss can differ in downstream performance \citep{liu2022pretraininglossbetterdownstream}.
Pretrained language models have a low intrinsic dimension for FT, which helps explain the
data efficiency of FT \citep{aghajanyan2020intrinsicdimensionalityexplainseffectiveness}.
In controlled settings, whether FT can extract a fact depends on how
PT stored it \citep{allen2023physics}, and reinforcement-learning post-training yields
additional capabilities only when PT leaves capacity for improvement
\citep{zhang2025interplaypretrainingmidtrainingrl}.
Mechanistic studies find that FT often adds a small modification to pretrained
capabilities \citep{jain2023mechanistically}, enhances mechanisms that PT already built
rather than creating new ones \citep{prakash2024finetuningenhancesexistingmechanisms}, and, in
reasoning FT, repurposes a direction already present in the base model's activations
\citep{ward2025reasoningfinetuningrepurposeslatentrepresentations}.
In LLMs, facts unknown to the base model are learned slowly by FT and, once learned,
increase hallucination \citep{gekhman2024doesfinetuningllmsnew}, and training on a teacher's
outputs transmits its traits through semantically unrelated data only when teacher and student
share a base model \citep{cloud2025subliminallearninglanguagemodels}.
Multi-task PT as a route to general representations
dates to \citet{caruana1997multitask} and \citet{aghajanyan2021muppetmassivemultitaskrepresentations}.
Meta-learning optimizes an initialization explicitly for fast adaptation
\citep{finn2017modelagnosticmetalearningfastadaptation,nichol2018firstordermetalearningalgorithms},
whereas standard PT optimizes only its own loss.

\textbf{Fine-tuning.} Models fine-tuned on scattered observations can infer and verbalize the latent structure behind
them \citep{treutlein2024connectingdotsllmsinfer}, and much of this out-of-context generalization
amounts to adding a steering vector toward a concept the pretrained model already represents
\citep{wang2025simplemechanisticexplanationsoutofcontext}.
Narrow FT on insecure code produces broad misalignment
\citep{betley2025emergentmisalignmentnarrowfinetuning}, and small amounts of FT in narrow
contexts can shift behavior far outside them \citep{betley2025weirdgeneralizationinductivebackdoors}, while FT on ``A is B''
does not teach ``B is A'' \citep{berglund2024reversalcursellmstrained}.
Model diffing studies how FT changes a model's representations, for example with
crosscoders that compare base and chat models
\citep{minder2025overcomingsparsityartifactscrosscoders}, and ablating undesired concept directions
during FT can steer how the fine-tuned model generalizes
\citep{casademunt2025steeringoutofdistributiongeneralizationconcept}.

\textbf{World representations and synthetic setups.} Our setup extends the synthetic
world-modeling experiments of \citet{park2026convergent}, where transformers trained on relational
statements over entities form a linear representation of the world. We measure this
representation and transfer it between models.
Sequence models trained on synthetic data form internal world representations, such as the board
state of Othello \citep{li2023emergent}, and large language models represent space and time
\citep{gurnee2023language}, although a generative model can pass standard world-model diagnostics
while its implied world model is far less coherent \citep{vafa2024evaluatingworldmodelimplicit}.
\citet{alain2018understandingintermediatelayersusing} use linear probes to assess information
encoded in representations. \citet{li2016convergentlearningdifferentneural}
study whether different runs of one architecture converge to similar features.
Representational similarity across runs is measured by SVCCA and CKA
\citep{raghu2017svcca,kornblithcka} and by model stitching
\citep{lenc2015understanding,bansal2021revisiting}. \citet{nishi2024representation} show in a
synthetic setting that knowledge editing disrupts a structured representation; we observe
related changes during FT.

\textbf{Distillation.} Knowledge distillation transfers a teacher's behavior
\citep{hinton2015distilling,gou2021knowledge}, and feature distillation adds intermediate-layer
hints \citep{romero2015fitnets}. Our representation distillation uses an intermediate-layer
loss with a teacher pretrained at the same model size. We assess whether the recipient acquires
the teacher's ability to generalize after FT, a transfer that requires retaining the distillation
loss during FT.

\textbf{Forgetting and regularized fine-tuning.} Catastrophic forgetting
\citep{mccloskey1989catastrophic} is classically mitigated by weight-space regularization
\citep{Kirkpatrick_2017,li2018explicit} or replay.
Related regularizers act on representations instead of weights: they keep the fine-tuned model's features close to the pretrained model's \citep{jung2016lessforgettinglearningdeepneural} or stable under injected noise \citep{Hua_2025}.
In language models, forgetting has been measured under continual instruction tuning
\citep{luo2025empiricalstudycatastrophicforgetting} and explained in part as FT skewing the model's implicit
inference of the task \citep{kotha2024understandingcatastrophicforgettinglanguage}.
In our experiments, activation-space
regularization distinguishes runs that can learn the task using their pretrained representations
from those that fit it by substantially changing those representations (Sec.~\ref{sec:hidden}).
\citet{achille2019criticallearningperiodsdeep} show that data arriving late in training cannot
reshape representations; FT on cross-group instances in Sec.~\ref{sec:ortho} provides a controlled
example of this effect.

\section{Experimental setup}
\label{app:setup}

\subsection{Notation and Terminology}
\label{app:notation}
This section defines the terminology used throughout the paper.

\textbf{Seeds.} A PT run is determined by its random seed. We refer to runs by their \emph{tier} (super, good, boundary, or bad, classified by FT generalization) rather than by their seed numbers. Runs pretrained under the ortho strategy are labeled as ortho to distinguish them from their counterparts under the shared strategy.

\textbf{Tests.} \emph{Across-generalization} is measured by fine-tuning on within-group instances from two groups and then evaluating on cross-group combinations; achieving this requires a joint representation. \emph{FT generalization} refers to the across-generalization score. The \emph{null band} refers to the region where accuracy is no higher than $0.205$: chance is $0.06$, the optimal input-independent answer is $0.10$, and ortho models reach at most $0.20$. Therefore, runs higher than $0.205$ are considered to exhibit across-generalization, while those no higher than $0.205$ do not. \emph{Tiers} classify seeds based on the peak FT generalization reached during plain FT: super $\ge 0.40$, good $0.30$--$0.40$, boundary $0.205$--$0.30$, and bad $\le 0.205$ ($3$, $12$, $47$, and $216$ runs, respectively). In the screening analysis, ``good runs'' refers to the top two tiers (super or good), totaling $15$ out of $278$ runs.

\textbf{Coordinate readout and answer readout.} The \emph{coordinate readout} is a linear mapping from the layer-5 representation to the ground-truth coordinates of entities (used for plotting maps and computing decodability). The \emph{answer readout} is a linear probe of the compositional answer obtained in a single forward pass at the \texttt{=} position (used as the step-0 predictor in Sec.~\ref{sec:reps}).

\textbf{Regularization strength.} In the synthetic models, the strength of the representation regularization term is $\lambda_{\mathrm{rep}} = 2.0$, applied at layer 5. In language model distillation, it ranges from $0.05$ to $1.0$ depending on the layer and model scale (App.~\ref{app:llm}).

\textbf{Scoring.} For the Pythia two-hop task in App.~\ref{app:llm}, we report the best single-evaluation accuracy. Unless otherwise noted, FT scores for synthetic models refer to FT generalization under the stated protocol, with regularized and plain values reported separately.

\textbf{Step-0 answer readout correlations.} The correlation of the answer readout is $r = 0.48$ with regularized in-distribution accuracy, $r = 0.64$ with regularized FT generalization, and $r = 0.32$ with plain FT generalization ($n = 278$ for all). Any claims regarding predicting FT outcomes using the answer readout refer to the regularized setting, as regularization measures learning when representation change is constrained (Sec.~\ref{sec:hidden}).

\subsection{Synthetic Models: Architecture and Training}
\label{app:training}
Unless otherwise stated, the synthetic model experiments (including the $278$ PT runs and the interventions performed on them) all use the following configuration. Only the PT seed differs across runs, as well as the FT seed where noted. The oracle experiments use a separate setup detailed in App.~\ref{app:oracle}.

\textbf{Architecture.} We use a $6$-layer Qwen2 transformer with a hidden size of $128$, a feed-forward size of $512$, and $4$ attention heads, with weights initialized at a standard deviation of $0.1$. Inputs use a character-level tokenizer (entities are represented by their ID strings, and no coordinates of world cities appear in the text) with a maximum sequence length of $256$.

\textbf{Pretraining data.} The corpus mixes two types of examples in equal proportions. Regular examples contain only relational statements about world cities. In-context examples first define one to four new entities with random coordinates, explicitly written as \texttt{p\_XXXX=(x,y)}, and then state relations using these entities; these entities are resampled in each example. An in-context ID may collide with a world city's ID, in which case the in-context coordinates override that city's coordinates within that example.

\textbf{Pretraining.} We use AdamW ($\beta_1{=}0.9$, $\beta_2{=}0.999$, $\epsilon{=}10^{-8}$), a learning rate of $6\times10^{-4}$, weight decay of $0.01$, gradient norm clipping at $1.0$, a linear schedule with a $50$-step warmup, a batch size of $128$, and bf16 precision. Models are trained on the relational corpus for $12$ epochs, corresponding to $656$k optimization steps.

\textbf{Fine-tuning.} FT loads only the pretrained weights and resets the optimizer and scheduler. The optimizer, weight decay, clipping, schedule, and batch size are all identical to PT, with a learning rate of $2\times10^{-4}$ and a $20$-step warmup, trained on the across-generalization task for approximately $33$k steps (the transfer test in App.~\ref{sec:xor} is about twice as long). Unless variation across FT seeds is reported, results use a single fixed FT seed. We evaluate periodically during training and report FT generalization under the stated protocol.

\textbf{Representation regularization.} At every step, the regularization term adds $\lambda \cdot \tfrac{1}{N}\sum_x \lVert h_5(x;\theta) - h_5(x;\theta_{\text{teacher}}) \rVert^2$ to the loss at layer $5$. The standard strength for synthetic models is $\lambda = 2.0$. The target is the model's own layer-$5$ representation, with two variants: \emph{fixed-target} regularization uses the pretrained representation as the target throughout the entire FT process (used in the strength sweep in Fig.~\ref{fig:probe}); \emph{periodic} regularization re-snapshots the target from the current model at the start of each FT segment (used for all-seed regularized results in Sec.~\ref{sec:reps} and Sec.~\ref{sec:hidden} as well as the language model experiments; App.~\ref{app:anchorabl} compares the two). Representation distillation (Sec.~\ref{sec:causal}) keeps the teacher checkpoint frozen and leaves the distillation loss active throughout FT. Fig.~\ref{fig:probe} varies $\lambda$ over $\{0,0.5,1,2,4,8,16,32,64\}$ to locate the interior optimum. Unless another strength is specified, regularization in synthetic models uses $\lambda = 2.0$.

\subsection{FT Generalization Metric}
\label{app:comps}
\begin{figure}[t]
\centering
\includegraphics[width=\textwidth]{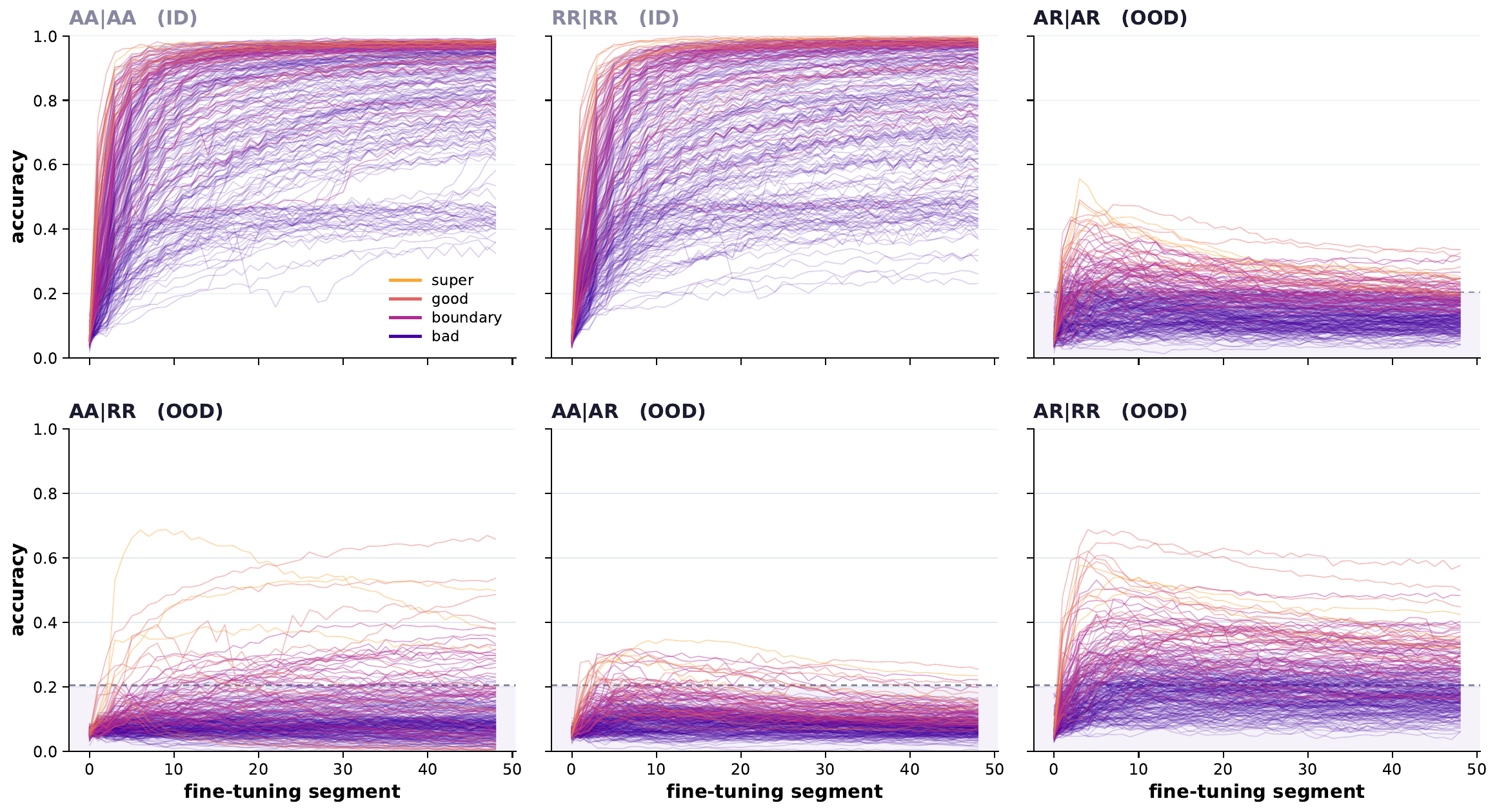}
\caption{\textbf{Fine-tuning curves by composition.} All $278$ runs, colored by
tier, for the two in-distribution compositions (top left) and the four
cross-group compositions whose mean is the paper's FT generalization. Every tier fits the task
in distribution; the tiers separate on each cross-group composition individually.}
\label{fig:app-comps}
\end{figure}

Let A denote the Americas and R denote the rest of the world. The \emph{composition} of an instance labels the group membership of its two segments. The two within-group compositions are AA$|$AA and RR$|$RR; the four cross-group compositions are AR$|$AR (both segments cross groups), AA$|$RR (the two segments lie in different groups), and the mixed types AA$|$AR and AR$|$RR. For the synthetic task, FT generalization is the average across the four cross-group compositions. Fig.~\ref{fig:app-comps} shows the FT trajectories for each composition; because FT is run in short segments with a full evaluation after each segment, the horizontal axis is measured in segments. The two within-group compositions are in-distribution, and all tiers (including bad runs) achieve high accuracy on them (final median of $0.89$). While the four cross-group compositions differ in absolute difficulty, the tier ordering is identical across all of them, confirming that the overall results are not driven by any single composition.

\textbf{Null band.} Under the $\pm10^{\circ}$ evaluation criterion, chance is $0.06$. The optimal input-independent strategy always answering the most frequent angle in the training distribution, achieves $0.10$. Ortho-pretrained models, which lack cross-group alignment by construction, achieve $0.05$--$0.20$ accuracy on cross-group compositions ($30$ runs, App.~\ref{app:xorseeds}). The upper bound of the null band, $0.205$, is set just above these three baselines, and only runs exceeding it are considered to exhibit across-generalization.

\subsection{Probing Procedure: Extraction, Decodability, and Projection}
\label{app:probe}
All maps, gallery panels, and coordinate decodability scores across all models and checkpoints use the same probing procedure.

\textbf{Extraction.} For each entity, we feed the prompt \texttt{angle(p\_XXXX,} (a PT task prefix followed by the entity ID and a comma, without the answer) into the frozen model and collect layer-5 hidden states at two token positions: the last digit of the ID and the trailing comma. Concatenating the two yields a $2 \times 128 = 256$-dimensional vector as the entity's representation. This procedure is applied to all $5{,}175$ entities while keeping the model frozen throughout; prior to any analysis, each dimension is standardized via z-score normalization.

\textbf{Decodability.} Linear regression is used to map representations to the entities' ground-truth coordinates $(x, y)$. Within-group decodability fits and evaluates this coordinate readout within the same entity group. Cross-group decodability tests whether a joint representation exists by fitting the coordinate readout on one group and evaluating it on the other, performed in both directions. We report the mean $R^{2}$ (the gallery grouping in App.~\ref{app:gallery} uses the more conservative minimum value). Joint representations achieve scores near or above $0$, whereas separated representations fall well below $0$ (the merged, partial, and disjoint groups in the gallery are separated by thresholds at $-3$ and $-10$). Because $R^{2}$ is normalized by coordinate variance, large negative values indicate that the cross-group prediction error is several times the coordinate variance.

\textbf{Projection for plotting.} Each displayed panel is a 2D projection whose basis is determined at the shown checkpoint. Specifically, we fit four coordinate readout directions (one $x$ and one $y$ for each group) and take the top two right singular vectors of this $4 \times 256$ matrix as the projection plane. After projection, the sign of each axis is determined by correlating it with true longitude, and the plane is rotated by $45^{\circ}$ for display. Entities are colored by true longitude, using one colormap per group. This projection is used strictly for visualization; coordinate decodability is computed using the full $256$-dimensional coordinate readout described above.

\subsection{Language-model experimental setup}
\label{app:llm}
\begin{table}[t]
\centering
\caption{\textbf{Pythia-410M distillation sweep.} Two-hop OOD accuracy (mean of four
FT seeds) after representation distillation from teacher A or teacher B into each recipient,
at each layer and weight, compared with the self-regularization control. Plain FT gives $0.278$ and $0.370$ for
the two recipients. Bold marks cells where every treated run exceeds every control run.}
\label{tab:p410}
\vspace{1em}
\footnotesize\begin{tabular}{llccc@{\hspace{1.6em}}ccc}
\hline
 & & \multicolumn{3}{c}{recipient 1 (plain $0.278$)} & \multicolumn{3}{c}{recipient 2 (plain $0.370$)} \\
layer & weight & teacher A & teacher B & control & teacher A & teacher B & control \\
\hline
16 & 0.05 & $0.402$ & $0.358$ & $0.355$ & $0.491$ & $0.444$ & $0.423$ \\
16 & 0.1  & $\mathbf{0.394}$ & $\mathbf{0.388}$ & $0.355$ & $0.455$ & $0.445$ & $0.467$ \\
16 & 0.2  & $0.394$ & $0.389$ & $0.370$ & $0.456$ & $0.484$ & $0.461$ \\
20 & 0.05 & $\mathbf{0.480}$ & $\mathbf{0.537}$ & $0.344$ & $0.487$ & $0.491$ & $0.367$ \\
20 & 0.1  & $\mathbf{0.467}$ & $\mathbf{0.539}$ & $0.381$ & $\mathbf{0.523}$ & $\mathbf{0.511}$ & $0.358$ \\
20 & 0.2  & $\mathbf{0.470}$ & $\mathbf{0.519}$ & $0.391$ & $\mathbf{0.512}$ & $\mathbf{0.519}$ & $0.314$ \\
22 & 0.05 & $\mathbf{0.511}$ & $\mathbf{0.519}$ & $0.397$ & $\mathbf{0.517}$ & $\mathbf{0.478}$ & $0.308$ \\
22 & 0.1  & $\mathbf{0.494}$ & $\mathbf{0.562}$ & $0.417$ & $\mathbf{0.542}$ & $\mathbf{0.497}$ & $0.361$ \\
22 & 0.2  & $\mathbf{0.486}$ & $\mathbf{0.541}$ & $0.389$ & $\mathbf{0.522}$ & $\mathbf{0.548}$ & $0.344$ \\
\hline
\end{tabular}
\end{table}
\subsubsection{Pythia-160M and Pythia-410M}
\label{app:llm-setup}
\textbf{Models.} We use Pythia-160M and Pythia-410M, along with their nine publicly released seed variants (\texttt{pythia-160m}, \texttt{pythia-160m-seed1} through \texttt{seed9}, and analogously for 410M). This provides ten PT runs per model scale that differ only in their random seed.

\textbf{Task.} \emph{relhop} is a synthetic two-hop relational task using fictional names: $400$ people, $60$ cities, and $20$ countries. Names are constructed from a $24$-syllable inventory so that every entity consists of multiple tokens and carries no prior semantic meaning. The corpus states the one-hop facts for every entity (``X lives in the city of Y'', ``the city of Y is in the country of Z'', each also presented in question form), but two-hop questions (``in which country does X live?'') are provided for only $240$ of the $400$ people. OOD evaluation poses two-hop questions about the remaining $160$ held-out people, requiring the model to compose stored one-hop facts. Chance accuracy is $0.05$, and the base models score $0$.

\textbf{Fine-tuning.} For both scales, FT runs for $24$ segments of $300$ steps each with a learning rate of $10^{-4}$. After each segment, we evaluate two-hop ID and OOD accuracy, Pile validation perplexity, and LAMBADA accuracy. Running plain FT with a single fixed FT seed across all ten PT seeds yields the ranges reported in Sec.~\ref{sec:llm} ($0.21$--$0.32$ for 160M, and $0.30$--$0.54$ for 410M).

Teachers are selected based on their scores under periodic regularization: using the auxiliary loss from Sec.~\ref{sec:causal} with $\lambda_{\mathrm{rep}} = 0.1$, targeting the model's own state and updated after each segment. Recipients are chosen from seeds with low plain FT scores. At 160M, the recipient is the lowest-scoring seed under plain FT. At 410M, the two recipients are the seed with the lowest score under both protocols ($0.30$ plain) and the seed with the second-lowest score under regularization ($0.39$ plain). The two teachers are the highest-scoring seeds under regularization ($0.48$ and $0.47$), the first of which scores only $0.30$ under plain FT.

\textbf{Distillation.} Representation distillation between Pythia seeds follows the synthetic model procedure: an auxiliary loss aligns the recipient's layer-$\ell$ states to the teacher's corresponding states on the task text. The teacher checkpoint remains frozen, and the distillation loss stays active throughout FT. The control group follows the exact same procedure, except the target is replaced with the recipient's own base states.

At 160M, we distill at layers $\ell \in \{2, 6, 10\}$ out of $12$, with a distillation weight of $0.5$ at layers 2 and 6, and $1.0$ at layer 10; we use four FT seeds per condition at layers 6 and 10, and two at layer 2.

At 410M, Table~\ref{tab:p410} sweeps layers $\ell \in \{16, 20, 22\}$ out of $24$ and distillation weights $\{0.05, 0.1, 0.2\}$ across every teacher--recipient pair, with four FT seeds per cell. Each cell is paired with a self-regularization control, and each recipient with a plain FT baseline (four seeds). A preliminary sweep was conducted earlier at layers 8 and 12 with weights $0.5$ and $2.0$ (three seeds per cell, using only the first teacher and first recipient). In that preliminary sweep, representation distillation did not outperform the control in any cell ($0.28$ vs.\ $0.40$ at layer 12, weight $2.0$), prompting the shift to deeper layers and lower weights in the table. We report the recipient's best two-hop OOD accuracy across segments.

\section{Additional experiments}
\label{app:experiments}

\subsection{Seed variance (Sec.~\ref{sec:lottery})}
\label{app:exp-seed}

\subsubsection{Pretraining Curves Across Seeds}
\label{app:curves}
\begin{figure}[t]
\centering
\includegraphics[width=\textwidth]{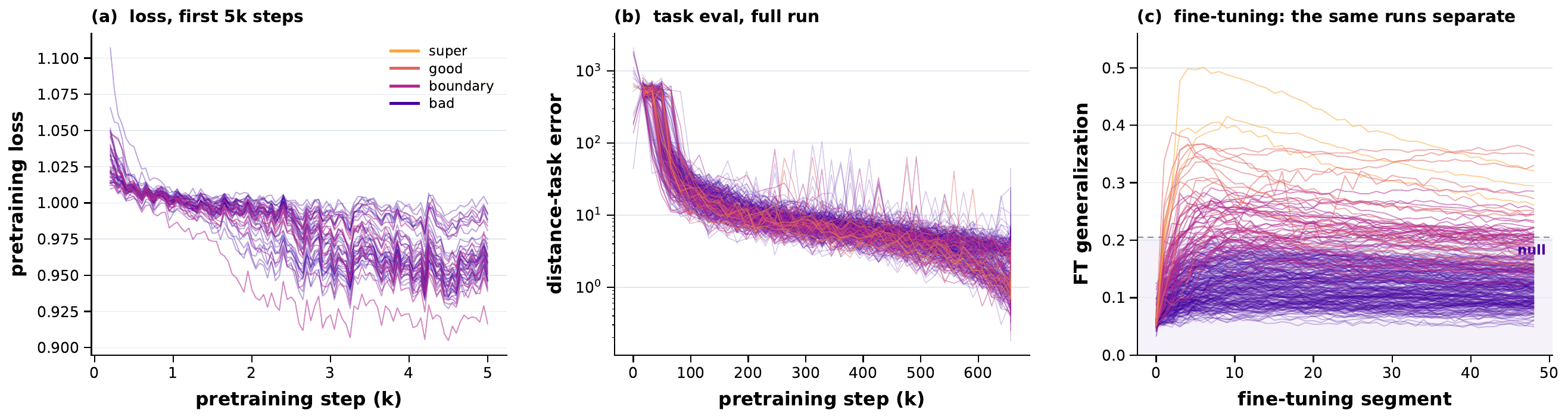}
\caption{\textbf{PT and FT trajectories by tier.} (a) PT loss over the first $5$k steps ($47$ sampled runs); (b) held-out error on the distance task across the full $656$k steps (all $278$ runs), both colored by final tier; neither metric separates the tiers. (c) FT generalization under the plain protocol for the same $278$ runs; the tiers separate early in FT and remain separated.}
\label{fig:app-curves}
\end{figure}

Fig.~\ref{fig:app-curves} compares the PT and FT trajectories across different seeds. Panel (a) shows the cross-entropy loss during the first $5$k steps for $47$ sampled runs, colored by their final tier. During the phase where loss changes the most rapidly, the tiers are indistinguishable. Panel (b) shows the held-out performance on the distance task across $42$ checkpoints over the entire $656$k steps for all $278$ runs; the remaining six tasks exhibit the exact same pattern. These evaluations fail to separate the tiers at any checkpoint. However, under the plain FT protocol in Sec.~\ref{sec:lottery}, these same $278$ runs separate by tier within the first few FT segments and remain separated throughout (panel (c)). Thus, differences in FT generalization are determined by PT, yet PT performance does not reveal these differences.

\begin{figure}[t]
\centering
\includegraphics[width=\textwidth]{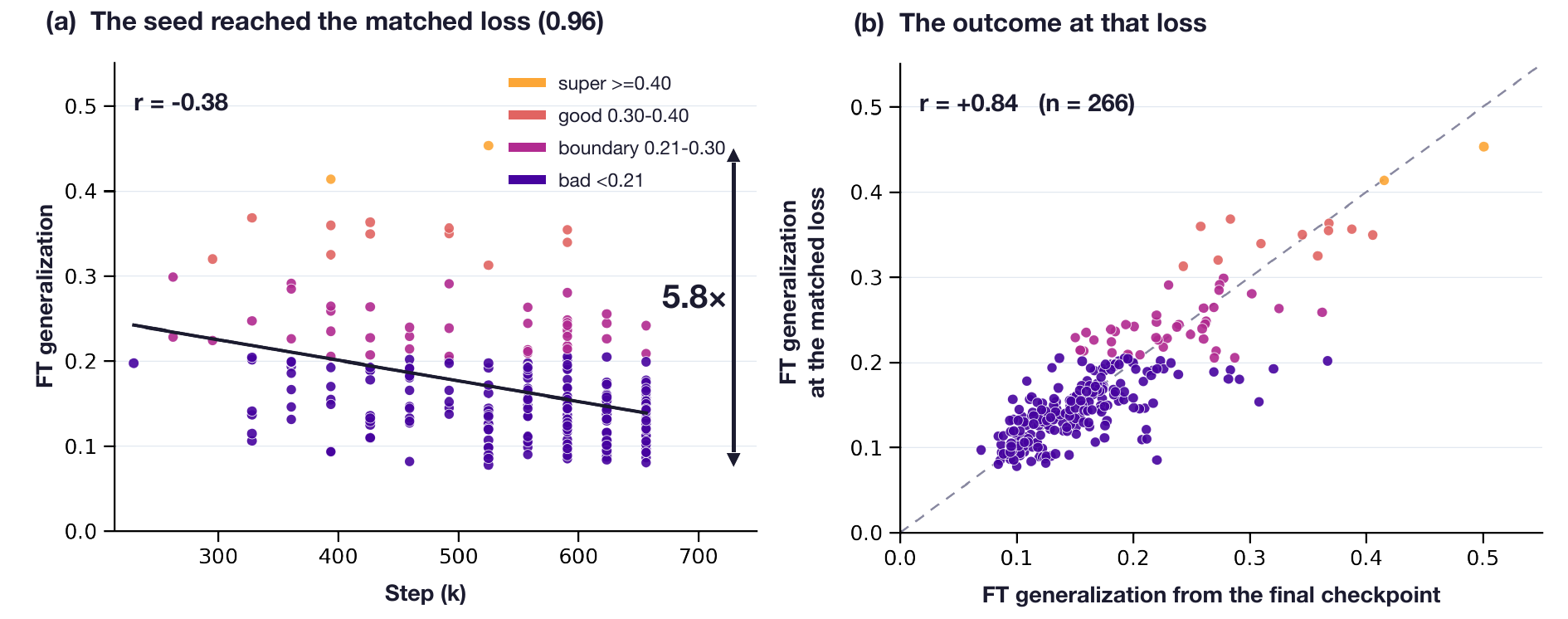}
\caption{\textbf{Matching PT loss does not remove the variation in FT generalization.}
(a) FT generalization for each of $266$ seeds, fine-tuned from its checkpoint at held-out
loss $0.960\pm0.005$, plotted against the step at which it reached that loss and colored by
eventual tier. (b) The same results plotted against FT generalization from each run's final
PT checkpoint.}
\label{fig:isoloss}
\end{figure}

Fig.~\ref{fig:isoloss} holds PT loss strictly fixed by construction rather than controlling for it approximately. For each of $266$ seeds, we take the checkpoint at which the held-out loss first reaches $0.960 \pm 0.005$ (a value every run passes through) and fine-tune from that checkpoint for half the standard budget ($24$ segments). The compute underlying these checkpoints still varies by $2.9\times$ (from $230$k to $656$k steps), yet FT generalization spans $0.078$--$0.453$ (median $0.151$), with $53$ of the $266$ runs exceeding the null band. Seeds that reach this matched loss earlier generalize slightly better (correlation with step count is $r = -0.38$). Each seed's outcome at the matched loss correlates with its outcome at the final checkpoint at $r = +0.84$ (Spearman $+0.81$). Hence, the ranking among seeds is already established by the time runs reach this loss, and subsequent training merely preserves this ranking.

\subsubsection{Variation Across Fine-Tuning Seeds}
\label{app:ftseed}
Table~\ref{tab:ftseed} reports the per-seed results from Fig.~\ref{fig:lottery}d: seven PT seeds, each paired with fourteen FT seeds. Super and good seeds never fall below $0.31$, bad seeds never exceed $0.10$, and only boundary seeds cross the null band. A one-way variance decomposition on this grid attributes $89.9\%$ of the outcome variance to the PT seed ($F(6, 91) = 125$). A replication on $43$ randomly sampled PT seeds $\times$ $10$ FT seeds attributes $73.6\%$ ($F(42, 387) = 29$). Thus, under both designs, the PT seed explains the vast majority of the variance (App.~\ref{app:stats}).

\begin{table}[t]
\centering
\caption{\textbf{Variation across FT seeds.} FT generalization across fourteen
FT seeds per PT seed, for seven seeds spanning the tiers.}
\label{tab:ftseed}
\vspace{1em}
\begin{tabular}{lccccc}
\hline
tier & $n$ & min & median & max & range \\
\hline
super & 14 & 0.336 & 0.392 & 0.446 & 0.110 \\
good & 14 & 0.315 & 0.342 & 0.355 & 0.040 \\
boundary & 14 & 0.169 & 0.228 & 0.347 & 0.178 \\
boundary & 14 & 0.139 & 0.231 & 0.308 & 0.168 \\
boundary & 14 & 0.093 & 0.129 & 0.308 & 0.214 \\
bad & 14 & 0.049 & 0.060 & 0.098 & 0.050 \\
bad & 14 & 0.066 & 0.084 & 0.098 & 0.032 \\
\hline
\end{tabular}
\end{table}

\subsubsection{Combinations of Initialization and Data Order Across Ten Seeds}
\label{app:grid}
\label{sec:pairing}
\begin{figure}[t]
\centering
\includegraphics[width=0.78\textwidth]{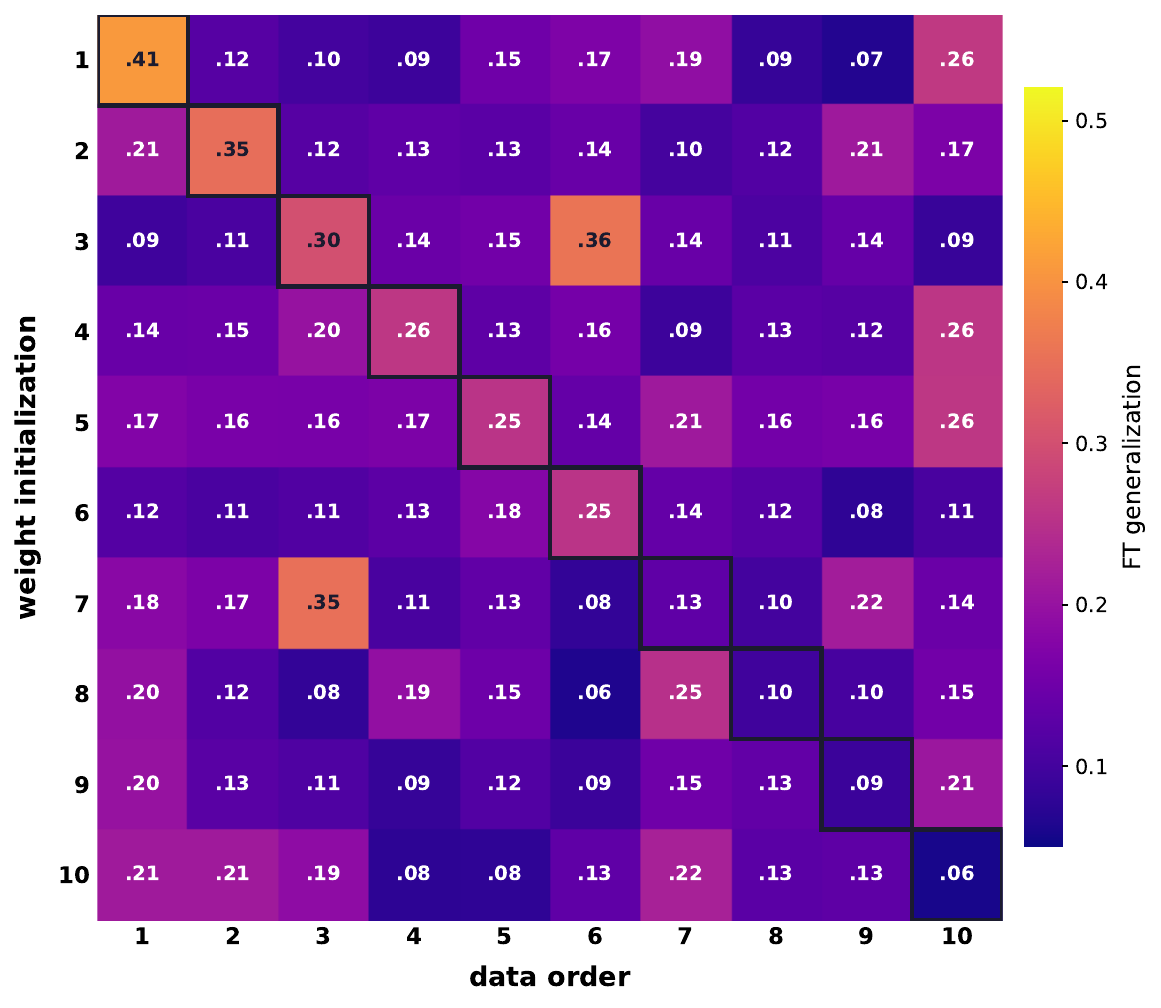}
\caption{\textbf{Interaction between initialization and data order across ten seeds.} Crossing the weight initializations and data orders of ten PT seeds; each cell reports FT generalization after full PT and plain FT, with boxed diagonal cells showing the original combinations. Rows and columns are seeds $1$--$10$, ordered by their original scores. Recombination scores cannot be predicted from initialization alone or from data order alone. The highest-scoring recombination ($0.36$) exceeds the original (diagonal) scores of both source seeds.}
\label{fig:grid100}
\end{figure}

The PT seed simultaneously determines weight initialization and data order. To test whether either factor alone can explain FT generalization, we vary them independently, following \citet{dodge2020finetuningpretrainedlanguagemodels} on decomposing FT seeds. Fig.~\ref{fig:grid100} crosses the initializations and data orders of ten seeds pairwise, yielding ninety recombinations, each pretrained in full and evaluated under plain FT. A seed's original score predicts neither how its initialization performs under other data orders nor how its data order performs with other initializations (correlations near zero). Thus, high FT generalization depends on the combination of initialization and data order, rather than on either component alone.

Fig.~\ref{fig:grid100} reports the full ten-seed grid. These seeds span all tiers, and all $100$ models (including the $90$ recombinations) use the plain FT protocol from Sec.~\ref{sec:lottery}. The diagonal shows each seed's original combination: three fall into the top two tiers ($0.30$--$0.41$), and four score poorly ($0.06$--$0.13$). Among the $90$ recombinations, fourteen exceed the null band and two reach the good tier. A seed's original score predicts neither the mean score of its initialization across other orders ($r = -0.12$) nor the mean score of its order across other initializations ($r = 0.02$). The seed with the highest original score averages $0.15$ when its initialization is paired with other orders, and $0.18$ when its order is paired with other initializations. The two highest-scoring recombinations ($0.36$ and $0.35$) both surpass the original scores of both of their source seeds ($0.30$ and $0.25$; and $0.13$ and $0.30$, respectively), so their performance cannot be attributed to either individual component. Each cell contains a single PT run, so we view this grid as evidence of an interaction between initialization and data order.

\subsection{How Representations Differ (Sec.~\ref{sec:reps})}
\label{app:exp-reps}

\subsubsection{Step-0 Answer Readout}
\label{app:signature}
\begin{figure}[t]
\centering
\includegraphics[width=\textwidth]{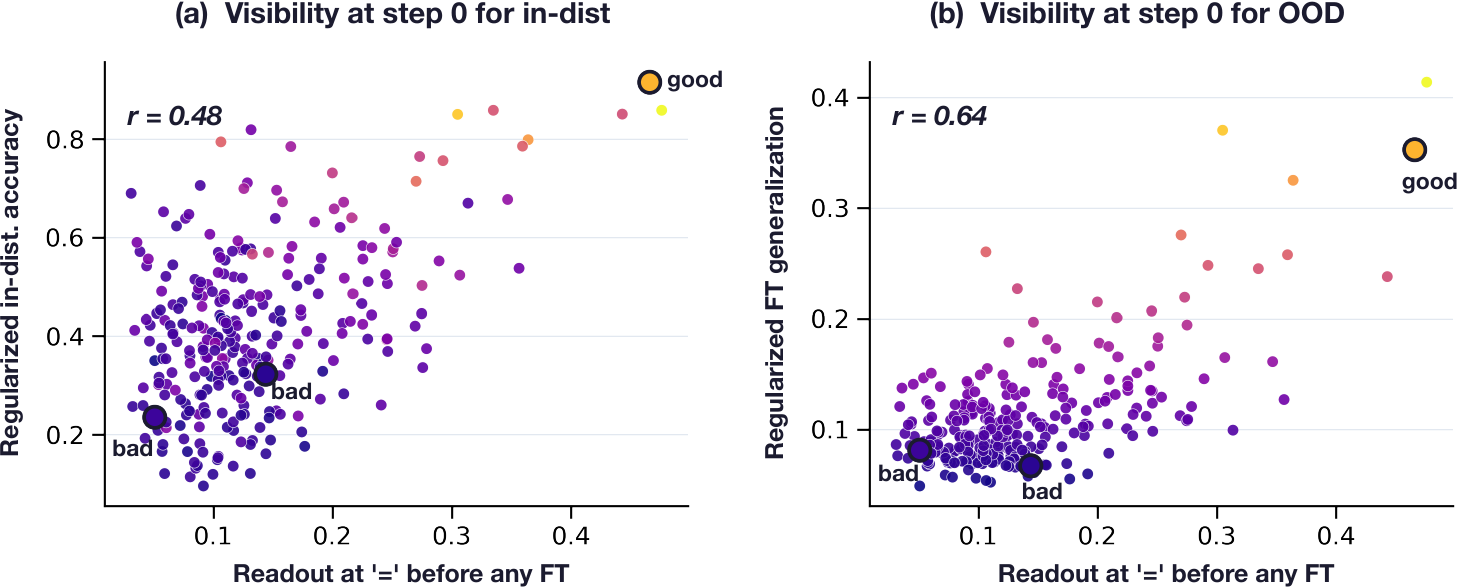}
\caption{\textbf{Step-0 answer readout.} (a, b) The step-0 answer readout at the \texttt{=} position of the pretrained checkpoint (method detailed in this appendix) predicts the outcomes achieved after regularized FT: in-distribution learning ($r = 0.48$) and FT generalization ($r = 0.64$; color indicates FT generalization, $n = 278$). Labels mark one good-tier seed and two bad-tier seeds.}
\label{fig:signature}
\end{figure}

The step-0 answer readout in Sec.~\ref{sec:reps} is a linear probe fitted on states obtained from a single forward pass of the frozen pretrained model. We take $2{,}500$ held-out \texttt{sangle} prompts, feed each prompt into the frozen model up to and including the \texttt{=} token, and collect the hidden state at that position (layer $6$ of $6$; layer $5$ gives similar results). Ridge regression ($\alpha = 5$) maps these states to the answer angle encoded as $(\cos \theta, \sin \theta)$; the regression is fitted solely on $80\%$ of the within-group prompts. The readout score is the circular correlation coefficient between the decoded angle and the true angle: the horizontal axis in Fig.~\ref{fig:signature}a--b is computed on held-out within-group prompts, and it is also computed separately on cross-group prompts. This probe measures the extent to which the task answer can be linearly decoded at the exact position where the model must output its answer. Because no gradient updates are applied to the model, this score characterizes the pretrained representation itself. Fig.~\ref{fig:signature}a--b shows that across seeds, it predicts both in-distribution learning and final FT generalization.

\subsubsection{Selecting Runs with Static Metrics}
\label{app:metrics}
\begin{figure}[t]
\centering
\includegraphics[width=0.92\textwidth]{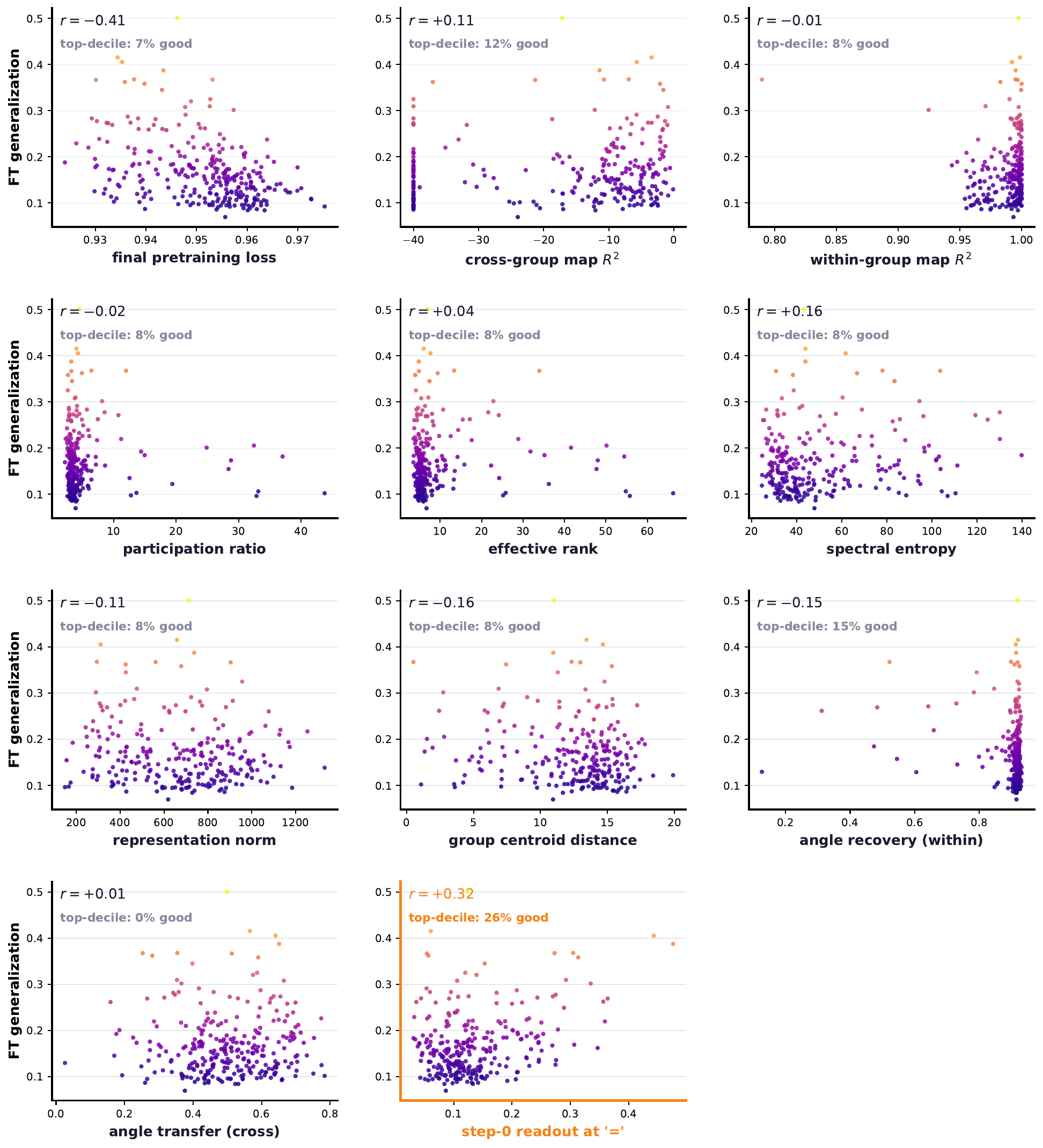}
\caption{\textbf{Static metrics are weak predictors of FT generalization.} Ten scalar summaries of
the pretrained model against eventual FT generalization across the seeds, each panel
annotated with its correlation and the fraction of good runs in its top decile (base rate
$5\%$). The step-0 answer readout (bottom right, orange) is the strongest selector.}
\label{fig:app-metrics}
\end{figure}

Fig.~\ref{fig:app-metrics} tests whether ten scalar metrics of the pretrained model can predict FT generalization across seeds, including:
\begin{itemize}
\item Final PT loss;
\item Several summary statistics of the final layer-5 representation: cross-group and within-group coordinate decodability, participation ratio, effective rank, spectral entropy, representation norm, and the distance between the centroids of the two groups;
\item Two angle recovery scores: a linear readout of each entity's within-group angular position, and the same readout applied across groups.
\end{itemize}
These metrics have limited discriminative power. Geometric summary statistics correlate with FT generalization at $|r| \le 0.16$, with good runs accounting for $8$--$12\%$ of their top decile against a base rate of $5\%$. PT loss reaches a correlation of $r = -0.41$, yet good runs make up only $7\%$ of its lowest-loss decile. By contrast, $26\%$ of the top decile of the step-0 answer readout (App.~\ref{app:signature}) are good runs. Its correlation with plain FT generalization in this figure is $r = 0.32$, and its correlation with regularized FT generalization is $r = 0.64$ (Fig.~\ref{fig:signature}). Therefore, what predicts FT generalization is whether the answer can be decoded from the representation, rather than the static geometric summary statistics tested here.

\begin{figure}[h]
    \centering
    \includegraphics[width=\textwidth]{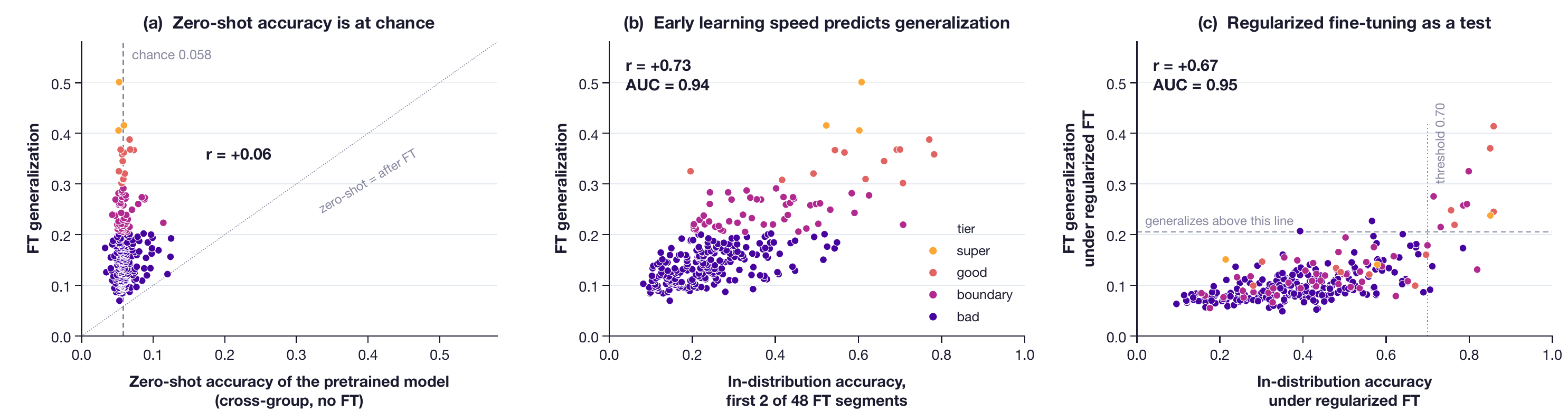}
    \caption{\textbf{Detecting runs that will generalize, without OOD data.} (a) Every pretrained model is at chance on the FT task before fine-tuning. (b) Accuracy after the first $4\%$ of FT already orders the seeds. (c) Under regularized FT, in-distribution accuracy with a fixed threshold of $0.70$ separates the seeds that generalize (AUC $0.95$).}
    \label{fig:app-predictors}
\end{figure}

Fig.~\ref{fig:app-predictors} adds three detection metrics that require no OOD data: all runs perform at chance level in zero-shot accuracy on the FT task, whereas early in-distribution accuracy under plain FT and in-distribution accuracy under regularized FT both distinguish runs that generalize.

\subsection{Oracle representations (Sec.~\ref{sec:oracle})}
\label{app:exp-oracle}

\subsubsection{Injection Procedure and Controls}
\label{app:oracle}
Fig.~\ref{fig:oracle} compares three conditions, each with $6$ PT seeds $\times$ $4$ FT seeds under the same protocol: no coordinate injection (Base), shuffled coordinate injection (Shuffle), and ground-truth coordinate injection (Oracle). Their final FT generalization scores are $0.17$, $0.17$, and $0.90$, respectively, while their in-distribution accuracies are $0.93$, $0.87$, and $0.97$.

\textbf{Procedure.} Each city's ground-truth coordinates are first normalized to $[-1, 1]^2$ in a fixed global coordinate system, then passed through a learnable linear projection $\mathbb{R}^2 \to \mathbb{R}^{128}$ and added to the residual stream at the city's final ID token. Injection occurs at the input of every decoder layer after the first layer, so layer $0$ receives only token inputs. Coordinates are injected throughout both PT and FT; entities defined in context receive their in-context coordinates. The Shuffle control applies a fixed random permutation to the world cities' coordinates. Consequently, each city receives a fixed but incorrect coordinate pair while sharing the exact same token stream as the oracle runs.

\textbf{Training protocol.} These runs use a separate standalone implementation with all sources of randomness seeded, over the same world of $5{,}175$ cities. It differs from the setup in Sec.~\ref{sec:setup} in the following ways:
\begin{itemize}
\item PT tasks: Eight tasks---angle, perimeter, triangle area, maximum and minimum pairwise distance, convex hull area and perimeter, and shortest open path;
\item Initialization scale: $0.02$;
\item PT schedule: Learning rate $3 \times 10^{-4}$ for $4$ epochs;
\item FT: Mixing $20\%$ PT replay into the FT data, trained for $40$ epochs.
\end{itemize}
All three conditions share this protocol, ensuring a controlled comparison across them.

\textbf{Test-time removal control.} Oracle models read coordinates directly from the injected channel starting at the first layer that receives the injection, rather than from the city tokens. Removing the injection at test time drops performance to chance or below, confirming that the model actually relies on the externally provided representation. Thus, the oracle establishes an upper bound on FT generalization when representations are provided externally.

\subsection{Representation distillation (Sec.~\ref{sec:distill})}
\label{app:exp-distill}

\subsubsection{Representation distillation ablations}
\label{app:routes}

Table~\ref{tab:knockout} removes components of the pipeline one by one on two bad-tier
recipients using the same initial distillation protocol. Without initial distillation,
regularized FT remains low ($0.09$). On one of the recipients, initial distillation followed
by plain FT reaches $0.20$ before dropping to $0.13$. Regularizing toward the
post-distillation state as a fixed target yields $0.12$ to $0.15$, and periodic
regularization yields $0.16$ to $0.17$. Retaining the distillation loss toward the teacher
representation during FT achieves the highest scores on both recipients ($0.28$ and
$0.27$), whereas an equal number of additional PT steps without distillation yields only
$0.10$. These comparisons show that both the initial distillation and the choice of target
during FT contribute to the observed gains.

\begin{table}[t]
\centering
\caption{\textbf{Distillation stage ablations.} FT generalization for variants of
the two-stage pipeline on two bad-tier recipients using the same initial distillation protocol.
The full-pipeline row is the matched grid of Fig.~\ref{fig:distill}b (mean of four FT seeds);
the other rows are single runs under the same $24$-segment protocol.}
\label{tab:knockout}
\vspace{1em}
\begin{tabular}{lcc}
\hline
pipeline variant & recipient A & recipient B \\
\hline
plain FT (no intervention) & $0.09$ & $0.07$ \\
matched additional PT, no distillation & $0.10$ & --- \\
regularized FT only, no initial distillation & $0.09$ & $0.10$ \\
initial distillation, then plain FT & $0.20 \to 0.13$ & $0.10$ \\
distillation $+$ fixed-target regularization & $0.15$ & $0.12$ \\
distillation $+$ periodic regularization & $0.16$ & $0.17$ \\
distillation $+$ teacher target (full pipeline) & $\mathbf{0.28}$ & $\mathbf{0.27}$ \\
\hline
\end{tabular}
\end{table}

\subsubsection{Distillation across all seeds}
\label{app:transplant}
\begin{figure}[t]
\centering
\begin{minipage}[t]{0.48\textwidth}
\centering
\includegraphics[width=\linewidth]{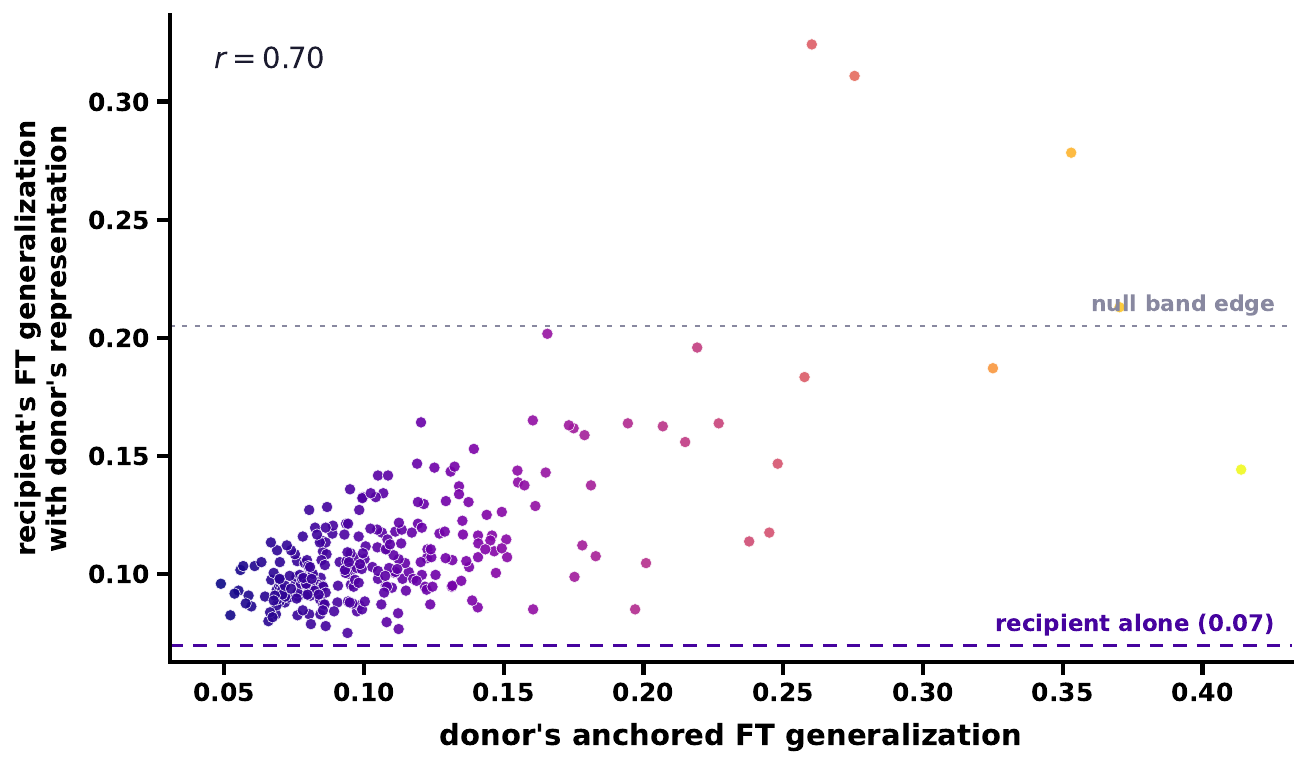}
\caption{\textbf{Distillation across all seeds.} Each teacher's representation is
distilled into the same bad-tier recipient using a fixed protocol. The recipient's FT
generalization correlates with the teacher's regularized FT generalization ($r=0.70$;
color: teacher score). The highest-performing teachers raise the recipient above the
null band.}
\label{fig:app-transplant}
\end{minipage}\hfill
\begin{minipage}[t]{0.48\textwidth}
\centering
\includegraphics[width=\linewidth]{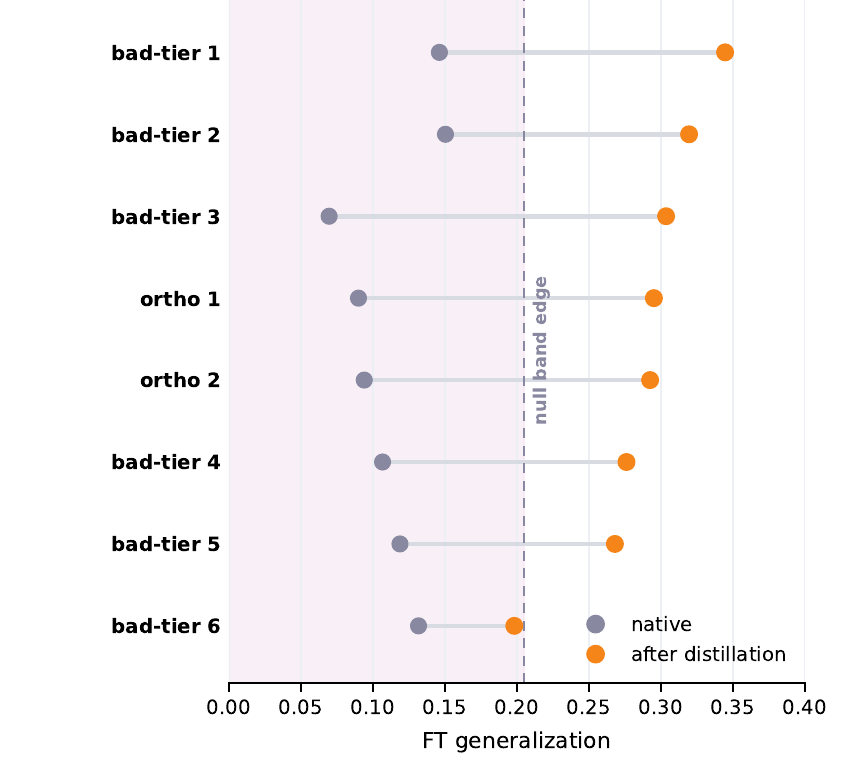}
\caption{\textbf{One teacher, many recipients.} The full two-stage pipeline distills one good-tier teacher into eight recipients; grey shows
baseline FT generalization and orange shows FT generalization after distillation. Seven of
eight recipients exceed the null band.}
\label{fig:app-templates}
\end{minipage}
\end{figure}

Fig.~\ref{fig:app-transplant} distills the representations of each of $274$ teachers into
the same bad-tier recipient, whose baseline FT generalization is $0.07$. All teachers are
distilled using the same fixed protocol: a short initial distillation phase followed by FT
with the distillation loss. Every teacher improves the recipient's score, and the magnitude
of the gain correlates with the teacher's regularized FT generalization ($r = 0.70$ on the
linear scale in Fig.~\ref{fig:app-transplant}; $0.65$ when taking the log of the scores as
in Fig.~\ref{fig:distill}c). The highest-scoring teachers lift the recipient above the null
band, reaching a maximum of $0.32$ under this short protocol; the full two-stage pipeline in
Sec.~\ref{sec:causal} uses a longer initial distillation phase and a tuned second stage.

Fig.~\ref{fig:app-templates} applies the full two-stage pipeline to distill from a good-tier
teacher into eight recipients: six bad-tier seeds and two ortho models. Seven of them
improve from baselines of $0.07$--$0.15$ to $0.27$--$0.35$---above the null band---while
the eighth reaches the upper edge of the null band ($0.20$). In a shorter protocol
distilling each of $29$ teachers into three bad-tier recipients ($50$k steps of initial
distillation and $16$ regularized segments), teacher rankings are consistent across
recipients (pairwise Spearman $\rho = 0.43$ to $0.61$), and the gains track the teacher's
regularized FT generalization (consistent with Fig.~\ref{fig:app-transplant}) rather than
its plain FT generalization ($r = -0.21$), because plain FT outcomes also reflect how much
FT reshapes the representation (Sec.~\ref{sec:hidden}).

\subsection{The ortho control (Sec.~\ref{sec:ortho})}
\label{app:exp-ortho}

\subsubsection{Cross-group generalization across PT strategies}
\label{sec:xor}
Under the same FT protocol, we compare the ortho control from Sec.~\ref{sec:ortho} against
two additional PT strategies: shared and weak. The shared model develops a joint
representation and achieves $0.40$ cross-group generalization, whereas the ortho model
maintains separate representations and reaches only $0.09$. The weak model adds cross-group
training after ortho PT; its representations become less sharply separated, yet remain
distinct, and cross-group generalization stays low ($0.11$; $0$ of $25$ runs exceed the null
band). The FT generalization of ortho consistently stays within the null band
($0.05$--$0.20$ across seeds and FT protocols), whereas shared runs span the full range
observed in Sec.~\ref{sec:lottery}. App.~\ref{app:ident} discusses why within-group data
underdetermine cross-group relationships, and App.~\ref{app:xorseeds} repeats this test
across different PT base models and FT seeds.

\subsubsection{Replication across seeds}
\label{app:xorseeds}
Table~\ref{tab:xorseeds} summarizes the replication results by PT strategy. None of the
ortho or weak runs exceed the null band on cross-group generalization ($0/30$ and $0/25$,
respectively; exact $95\%$ upper confidence bounds on the true rate are $10\%$ and $11\%$),
whereas the FT generalization of shared varies across seeds as described in
Sec.~\ref{sec:lottery}. Thus, adding cross-group data after ortho PT does not produce
cross-group generalization. Replaying general data at a ratio of up to $0.5$ during FT also
does not alter these results. This separation is a property of the PT strategy itself
rather than a quirk of any specific run.

\begin{table}[t]
\centering
\caption{\textbf{Across-generalization across seeds.}
Across replications by PT policy, no ortho or weak run exceeds the null band (0.20) on
across-generalization.}
\label{tab:xorseeds}
\vspace{1em}
\begin{tabular}{llccc}
\hline
policy & test & runs & range & above the null band \\
\hline
shared & across-generalization & 278 & $0.07$--$0.50$ & $62$ (the lottery, Sec.~\ref{sec:lottery}) \\
ortho & across-generalization & 30 & $0.05$--$0.20$ & $0$ \\
weak & across-generalization & 25 & $0.06$--$0.19$ & $0$ \\
\hline
\end{tabular}
\vspace{1em}

\raggedright \footnotesize All rows use the full-length protocol, with five bases
$\times$ five FT seeds per cell (ortho across-generalization includes five earlier runs).\par
\end{table}

\subsubsection{Identifiability of cross-group relations}
\label{app:ident}
The ortho corpus constrains within-group structure without directly specifying cross-group
relations, leaving the relative alignment between the two groups underdetermined. However,
some PT tasks (including Manhattan distance) as well as the FT task (\emph{signed} angle) are
not invariant to rotating or reflecting a single group independently; these tasks restrict
the admissible transformations, and a formal identifiability statement should characterize
the remaining ambiguity. This argument describes the indeterminacy left by the available
data, rather than proving impossibility.
Empirically, all $30$ ortho replicates remain at or below the null band, where the
input-independent marginal answer optimum is approximately $0.10$. The weak strategy
introduces cross-group data after PT, providing information about the relative alignment,
yet none of its runs exceed the null band ($0$ out of $25$; see App.~\ref{sec:xor}).

\subsection{Representation regularization (Sec.~\ref{sec:hidden})}
\label{app:exp-holding}

\subsubsection{Reshaping and forgetting under fixed-target regularization}
\label{app:reshape}
\begin{figure}[t]
\centering
\includegraphics[width=\textwidth]{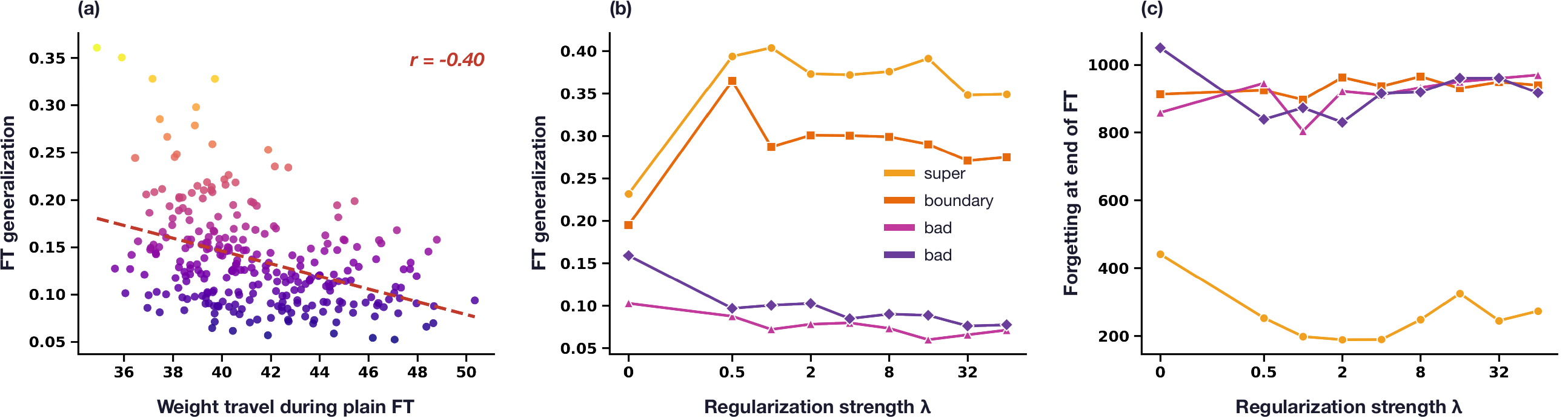}
\caption{\textbf{Parameter change and fixed-target regularization.}
(a) Across $276$ seeds, parameter change during plain FT is negatively correlated with FT
generalization ($r=-0.40$).
(b) Four seeds that end plain FT ($\lambda=0$) at $0.10$--$0.23$; by tier, which is assigned
by the peak reached under plain FT, they are one super, one boundary, and two bad seeds.
(c) Forgetting over the same sweep, measured as the held-out error on the PT distance task at
the end of FT (coordinate units). Without replay data, regularization reduces this error for
the super seed; the other three retain high error.}     
\label{fig:probe}
\end{figure}

Under plain FT, total parameter change correlates negatively with FT generalization
($r = -0.40$, Fig.~\ref{fig:probe}a). Differences across tiers are concentrated from the
embeddings through layer 5, whereas layer 6 and the output head change by similar amounts
across all tiers (App.~\ref{app:layers}). To test the effect of constraining these changes,
we select four seeds that end up in the boundary or bad tiers under plain FT (one super, one
boundary, and two bad seeds by tier; $0.10$ to $0.23$) and regularize their layer-5
representations toward their respective pretrained states with increasing strength
$\lambda$ (Fig.~\ref{fig:probe}b). The super and boundary seeds improve to $0.30$--$0.40$,
while the two bad seeds remain near chance. For the super seed, forgetting is also reduced:
held-out error on the PT distance task at the end of FT (in world-coordinate units) drops
from $435$ to $183$ at the tuned strength, whereas forgetting is unchanged for the other
three seeds (Fig.~\ref{fig:probe}c). Thus, fixed-target regularization can improve FT
generalization and reduce forgetting without replay data, though its effectiveness depends
on the pretrained representation; the all-seed regularized results in
Sec.~\ref{sec:hidden} use the periodic variant (App.~\ref{app:anchorabl}).

\subsubsection{Per-layer parameter change}
\label{app:layers}
\begin{figure}[t]
\centering
\includegraphics[width=0.62\textwidth]{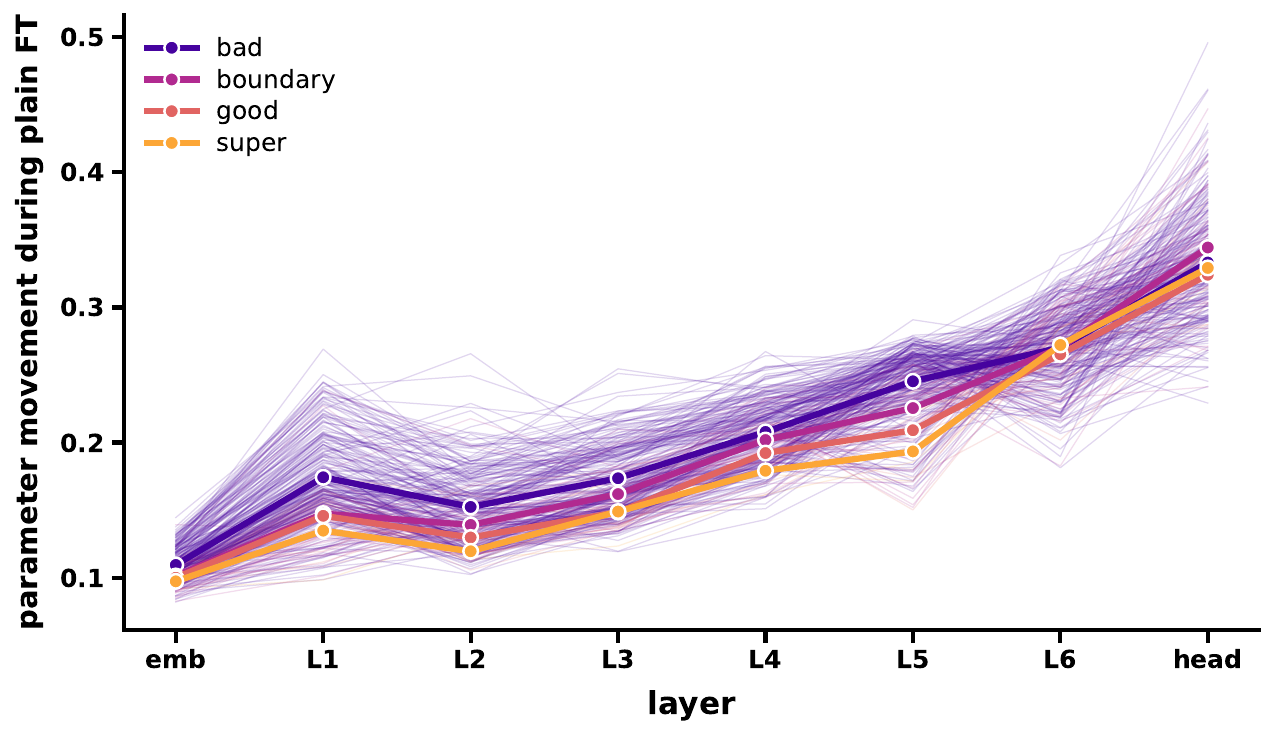}
\caption{\textbf{Per-layer parameter change.} Relative parameter change during plain FT
by layer, with thin lines for individual runs and bold lines for tier means. The tiers differ
in the embedding through layer 5 and show similar parameter changes at layer 6 and the output head.}
\label{fig:app-layers}
\end{figure}

Fig.~\ref{fig:app-layers} reports parameter change during plain FT
(Sec.~\ref{sec:dynamics}) by layer for $276$ of the $278$ runs, those with saved parameter
trajectories. Total parameter change correlates with FT generalization at $r=-0.40$ across
these runs and at $-0.30$ among the runs that fully fit the FT set, so the association is
not confined to runs that underfit the FT set. From the embedding through
layer 5, runs with high FT generalization show $5$--$15\%$ less parameter change than typical runs,
and bad-tier runs show the largest changes. Layer 6 and the output head show similar parameter changes across tiers.

\subsubsection{Plain FT, fixed-target regularization, and periodic regularization}
\label{app:anchorabl}
\begin{figure}[t]
\centering
\includegraphics[width=\textwidth]{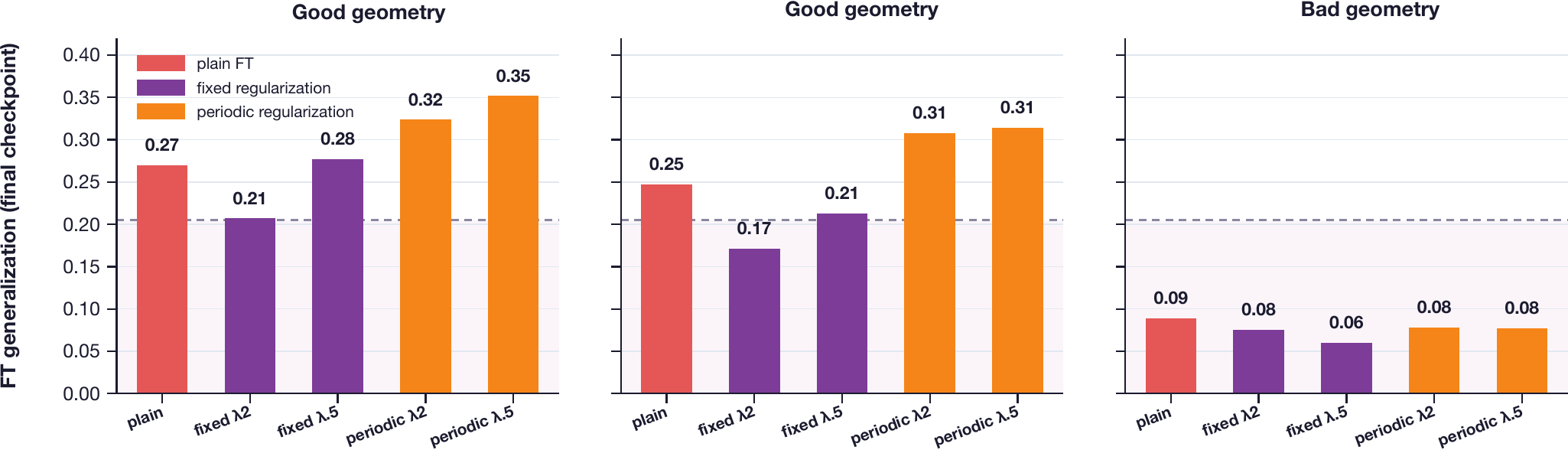}
\caption{\textbf{Plain FT, fixed-target, and periodic regularization.} Final-checkpoint FT
generalization under plain FT, fixed-target regularization, and periodic regularization at two
strengths. For the good-tier seeds, periodic regularization gives the highest final scores,
while strong fixed-target regularization lowers both seeds' tiers. The bad-tier seed remains
within its tier under all conditions.}
\label{fig:app-anchorabl}
\end{figure}

Fig.~\ref{fig:app-anchorabl} compares plain FT, fixed-target regularization, and periodic
regularization in terms of FT generalization at the final checkpoint under the same
$48$-segment protocol. Fixed-target regularization targets the pretrained representation
throughout the entire FT process, whereas periodic regularization updates the target using
the current model at each segment; we evaluate both methods at two strengths each on two
good-tier seeds and one bad-tier seed.
For the good-tier seeds, periodic regularization achieves the highest final scores ($0.31$
to $0.35$). Plain FT reaches $0.25$ to $0.27$, which is below its earlier peak; fixed-target
regularization reaches $0.17$ to $0.28$, and under the stronger fixed-target setting, both
good-tier seeds drop below their original tier. The bad-tier seed remains at $0.06$ to
$0.09$ across all conditions (Sec.~\ref{sec:dynamics}). These results indicate that updating
the target periodically preserves FT generalization better than anchoring the
representation near its pretrained state.

\subsection{Statistical notes (Secs.~\ref{sec:lottery}--\ref{sec:hidden})}
\label{app:stats}
This section reports uncertainty estimates for the paper's main quantitative claims, all
computed from the runs described above with no additional training.

\begin{figure}[t]
\centering
\includegraphics[width=0.92\textwidth]{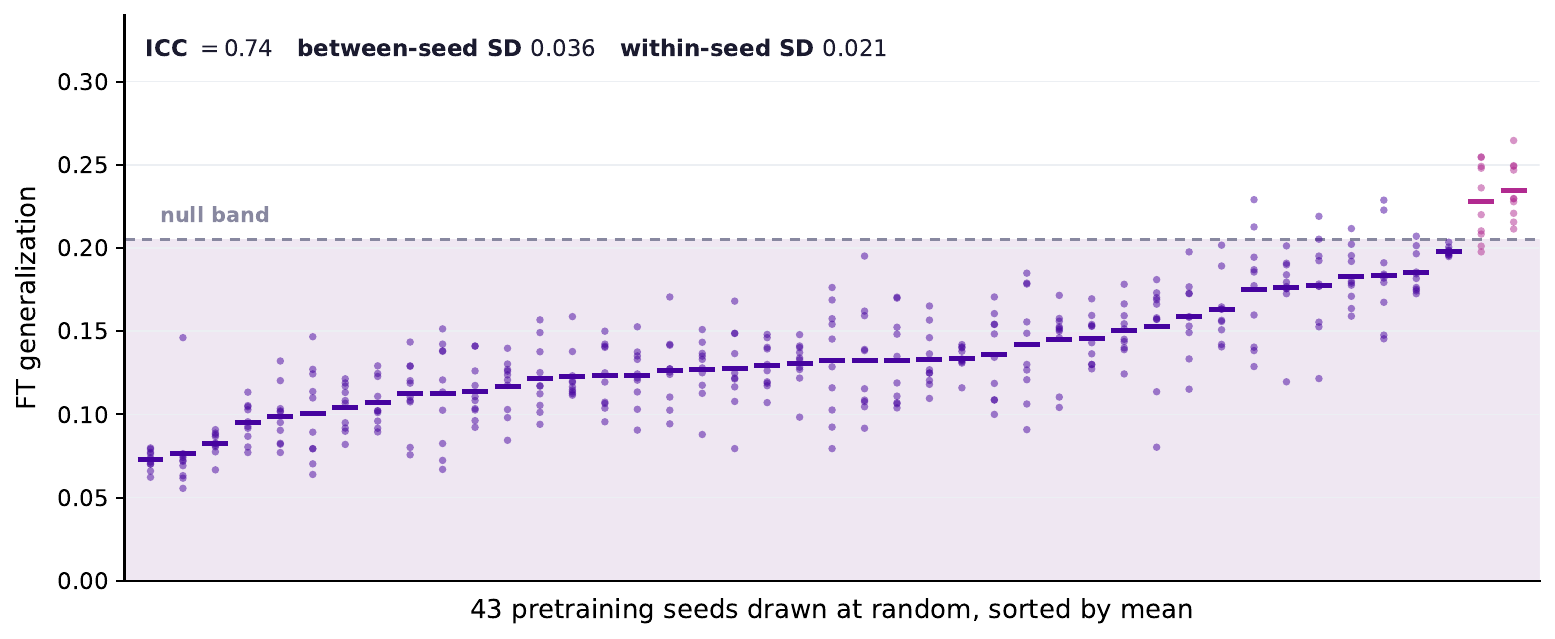}
\caption{\textbf{Variation across randomly sampled PT seeds.} Forty-three PT seeds,
each fine-tuned under ten FT seeds (dots), sorted by their mean FT generalization (bar) and
colored by the mean's tier. The between-seed standard deviation ($0.036$) exceeds the
within-seed standard deviation ($0.021$), with an intraclass correlation of $0.74$.}
\label{fig:icc}
\end{figure}

\textbf{Variation across PT seeds versus FT seeds.} Treating the $7 \times 14$ grid in
Table~\ref{tab:ftseed} as a one-way random-effects design, PT seeds explain $89.9\%$ of the
variance in FT generalization (intraclass correlation coefficient $0.899$;
$F(6, 91) = 125$, $p < 10^{-15}$). Because these seven seeds were selected to span all
tiers, we repeated this analysis on $43$ randomly sampled PT seeds, each fine-tuned with
$10$ FT seeds ($430$ runs total). In this sample, the intraclass correlation coefficient is
$0.736$ ($F(42, 387) = 28.9$, $p < 10^{-15}$); the between-seed standard deviation is
$0.036$, the within-seed (FT) standard deviation is $0.021$, and the median seed varies by
a range of $0.06$ across its ten FT seeds (Fig.~\ref{fig:icc}). In both samples, variation
across PT seeds exceeds variation across FT seeds.

\textbf{Screening by PT loss and answer readout.} Across all seeds ($n = 278$), final PT
loss correlates with FT generalization at $r = -0.41$ ($95\%$ confidence interval $-0.50$
to $-0.31$). In the lowest-loss decile, $2$ out of $27$ runs are high-performing runs
(high-performing $=$ super or good tier; base rate $5.4\%$; $p = 0.43$). In the highest
answer-readout decile, $7$ out of $27$ runs are high-performing
($p \approx 5 \times 10^{-4}$ relative to the base rate). Thus, the answer-readout decile
is significantly enriched relative to the base rate; however, a bootstrap confidence
interval for the direct difference between the two deciles ($95\%$ CI $-0.04$ to $0.41$)
does not distinguish the two screening methods, and these results complement the causal
interventions in Sec.~\ref{sec:causal}.

\textbf{In-distribution accuracy and FT generalization.} Under plain FT, in-distribution
accuracy correlates with FT generalization at $r = 0.51$ ($0.42$ to $0.59$); under
regularized FT, the correlation is $r = 0.67$ ($0.61$ to $0.73$). Because both correlations
are computed over the same $278$ runs, we compare them using a test for dependent,
non-overlapping correlations: the difference ($0.16$) is significant (Steiger $Z = 3.2$,
$p = 0.001$; paired bootstrap $95\%$ CI $0.08$ to $0.24$). Under regularized FT,
in-distribution accuracy remains informative because training does not force every run to
fit the FT dataset (Fig.~\ref{fig:hidden}).

\textbf{Exact binomial upper bounds.} For conditions with zero observed successes
(Table~\ref{tab:xorseeds}), we report exact binomial upper bounds: $0$ out of $30$ ortho
replicates exhibit cross-group generalization (rate less than $10\%$ at the $95\%$
confidence level), and $0$ out of $25$ weak runs do so (less than $11\%$).

\subsection{Behavioral distillation}
\label{app:behavdist}
Behavioral distillation replaces the layer-5 representation objective with
$\mathrm{KL}\!\left(p_{\theta}\,\|\,p_{\theta_{\mathrm{good}}}\right)$ at temperature
$\tau = 1$ over each ground-truth token. The corpus, step count, schedule, and FT protocol
are identical to those of representation distillation. Behavioral distillation reduces KL
divergence from the teacher and increases top-1 agreement, yet improvements in FT
generalization appear only under representation distillation.

\begin{table}[h]
\centering
\begin{tabular}{lccc}
\toprule
& s43 & sd816 & ortho \\
\midrule
KL to teacher, before $\to$ after & $0.060\to0.026$ & $0.095\to0.021$ & $0.275\to0.020$ \\
top-1 agreement, before $\to$ after & $0.81\to0.83$ & $0.81\to0.84$ & $0.76\to0.81$ \\
\midrule
FT generalization, plain FT & $0.096$ & $0.075$ & $0.080$ \\
FT generalization, behavior distillation & $0.073$ & $0.073$ & $0.078$ \\
FT generalization, representation distillation & $0.280$ & $0.271$ & $0.320$ \\
\midrule
FT generalization, behavior at $5\times$ weight & $0.110$ & $0.104$ & --- \\
\bottomrule
\end{tabular}
\end{table}

\begin{figure}[h]
    \centering
    \includegraphics[width=0.9\textwidth]{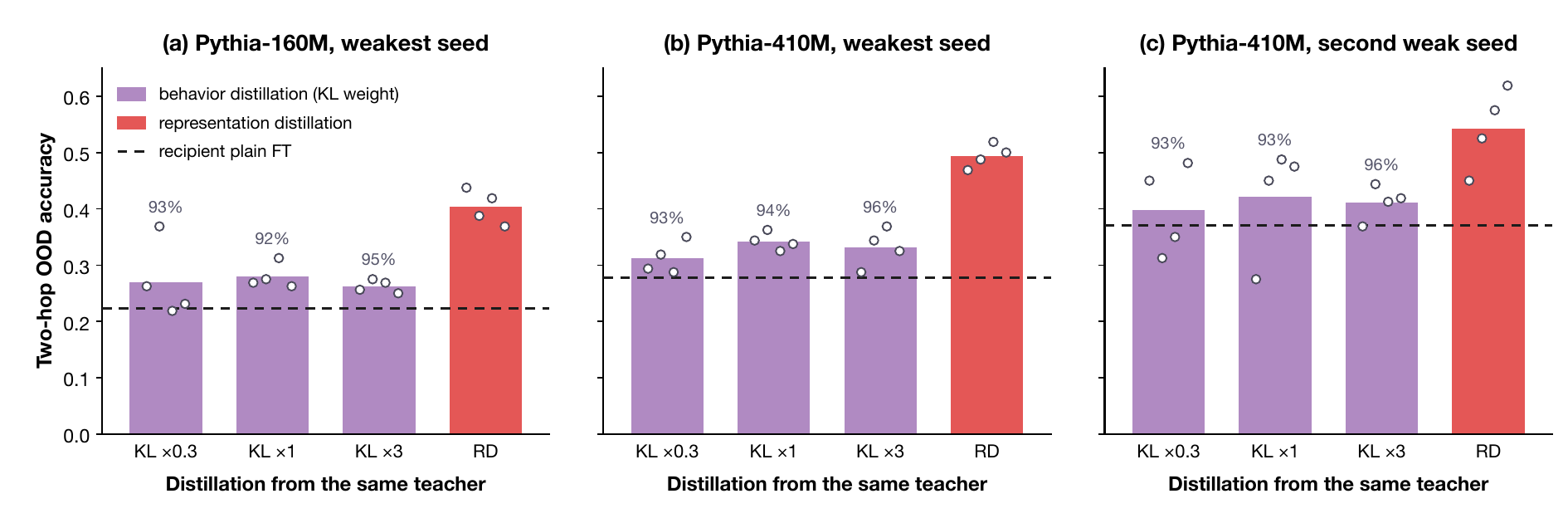}
    \caption{\textbf{Behavior distillation in Pythia.} Behavior distillation against representation distillation, with the same teacher. The percentage under each orange bar is the share of next tokens on which student and teacher agree at the end of fine-tuning.}
    \label{fig:app-behav-pythia}
\end{figure}

The loss is evaluated at $\tau=1$ over a $73$-token vocabulary.
Fig.~\ref{fig:app-behav-pythia} repeats the comparison in Pythia with the same teacher: behavior
distillation brings next-token agreement with the teacher to $92$--$96\%$ but leaves OOD accuracy
close to plain FT, while representation distillation raises it.

\section{Visualizations}
\label{app:viz}

\subsection{Gallery of representations across seeds}
\label{app:gallery}
\begin{figure}[t]
\centering
\includegraphics[width=\textwidth]{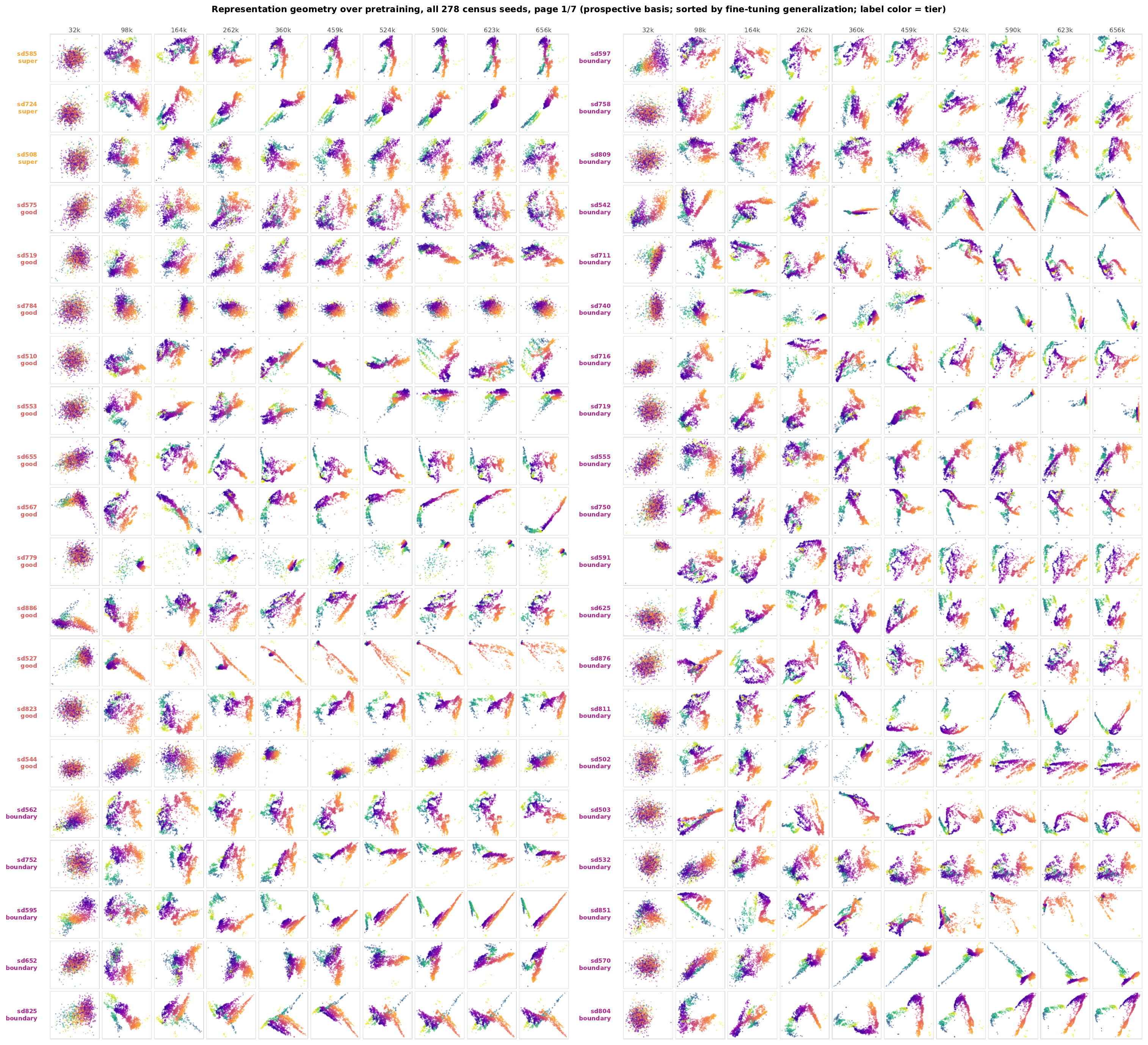}
\caption{\textbf{Representation maps across PT seeds.} Layer-5 representation maps
for seeds (rows) at ten PT checkpoints (columns), with a basis fitted separately at each
checkpoint. Distinct maps emerge early in PT and persist. Label colors indicate each seed's
FT generalization tier; static summaries of the representation are compared with tier in
App.~\ref{app:metrics}. Continued in Figs.~\ref{fig:gallery-p2}--\ref{fig:gallery-p7}.}
\label{fig:gallery-p1}
\end{figure}

\begin{figure}[p]
\centering
\includegraphics[width=\textwidth]{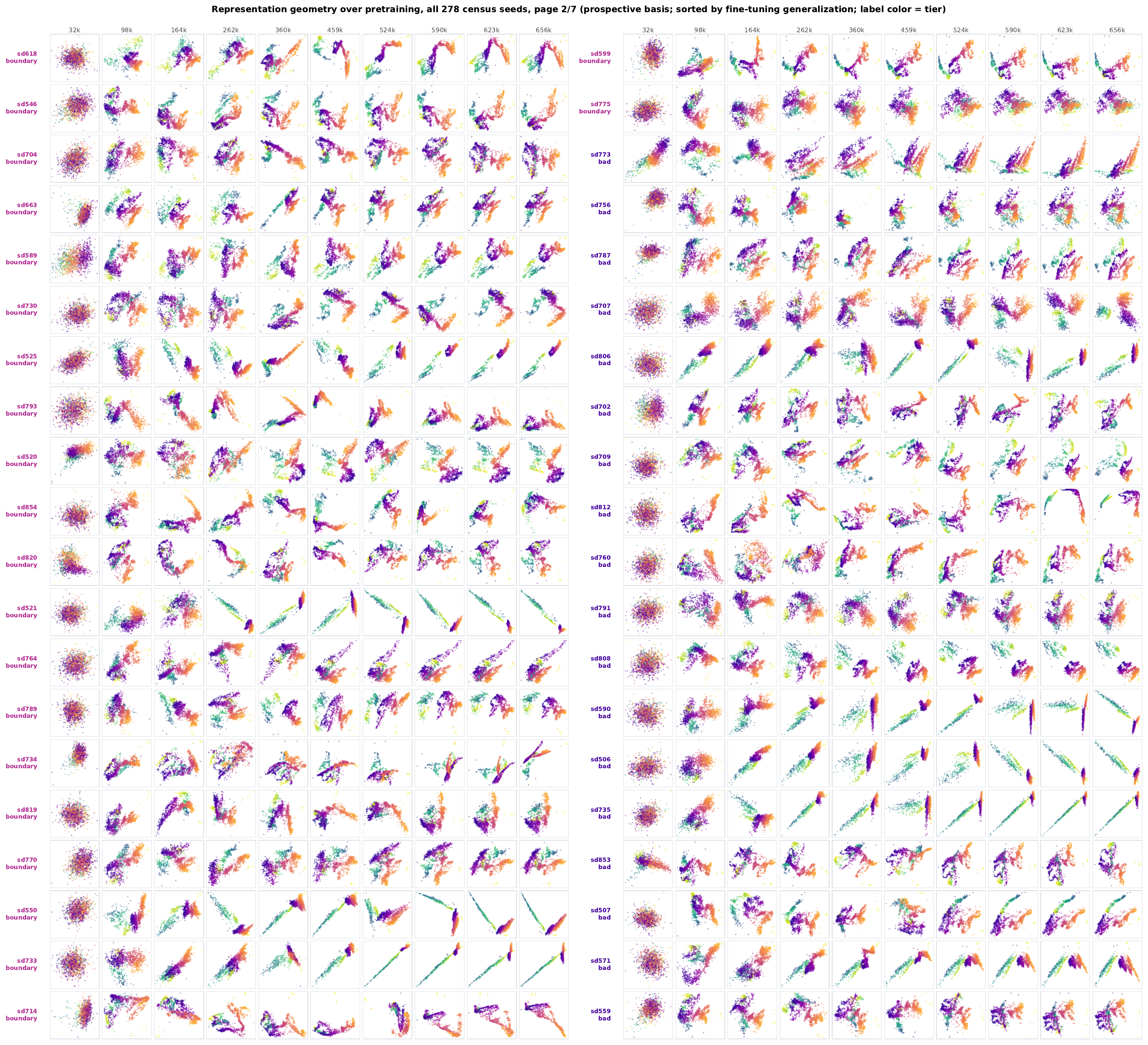}
\caption{\textbf{Representation maps across PT seeds (continued).} Additional seeds in the
layout of Fig.~\ref{fig:gallery-p1}.}
\label{fig:gallery-p2}
\end{figure}

\begin{figure}[p]
\centering
\includegraphics[width=\textwidth]{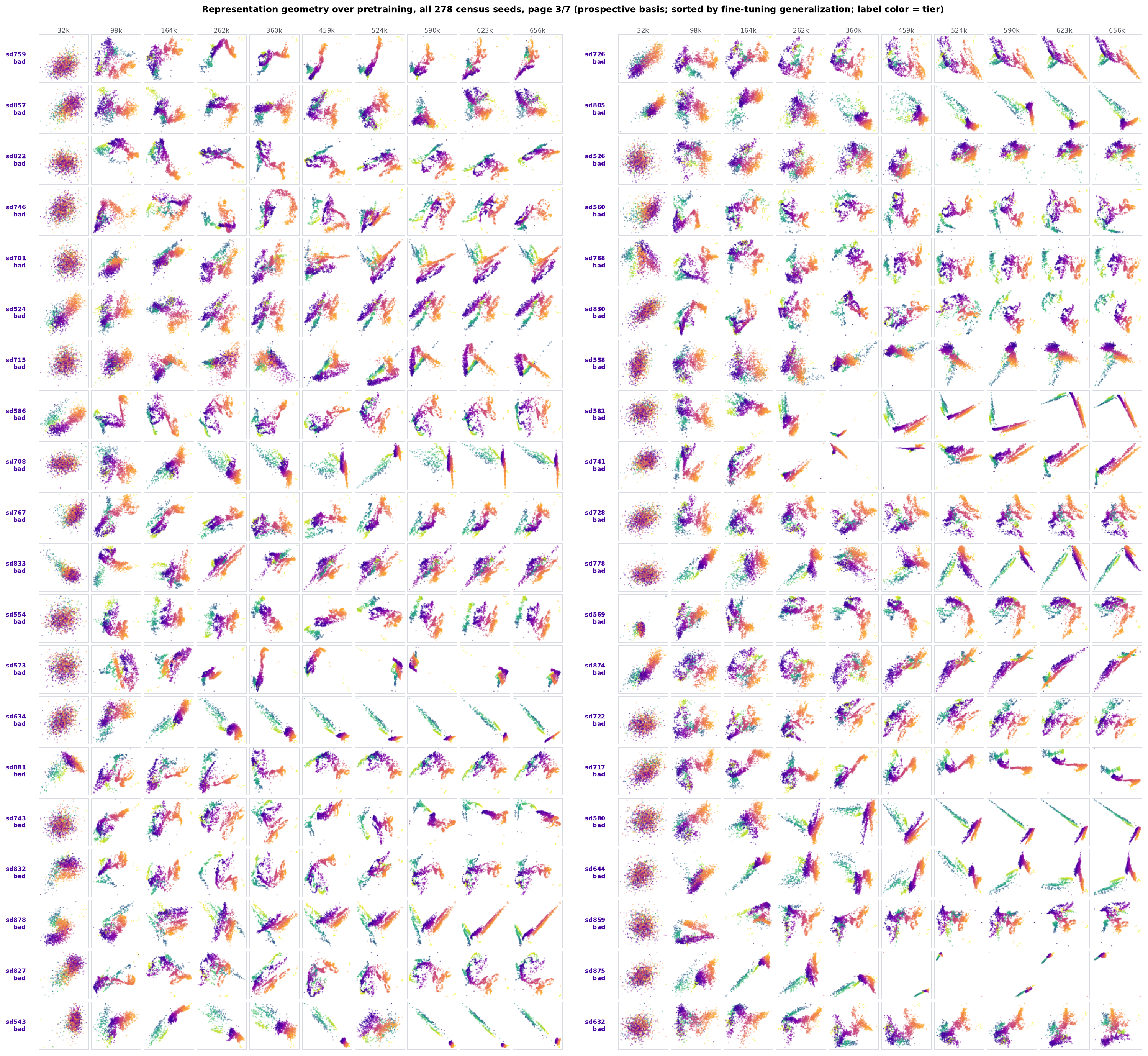}
\caption{\textbf{Representation maps across PT seeds (continued).} Additional seeds in the
layout of Fig.~\ref{fig:gallery-p1}.}
\label{fig:gallery-p3}
\end{figure}

\begin{figure}[p]
\centering
\includegraphics[width=\textwidth]{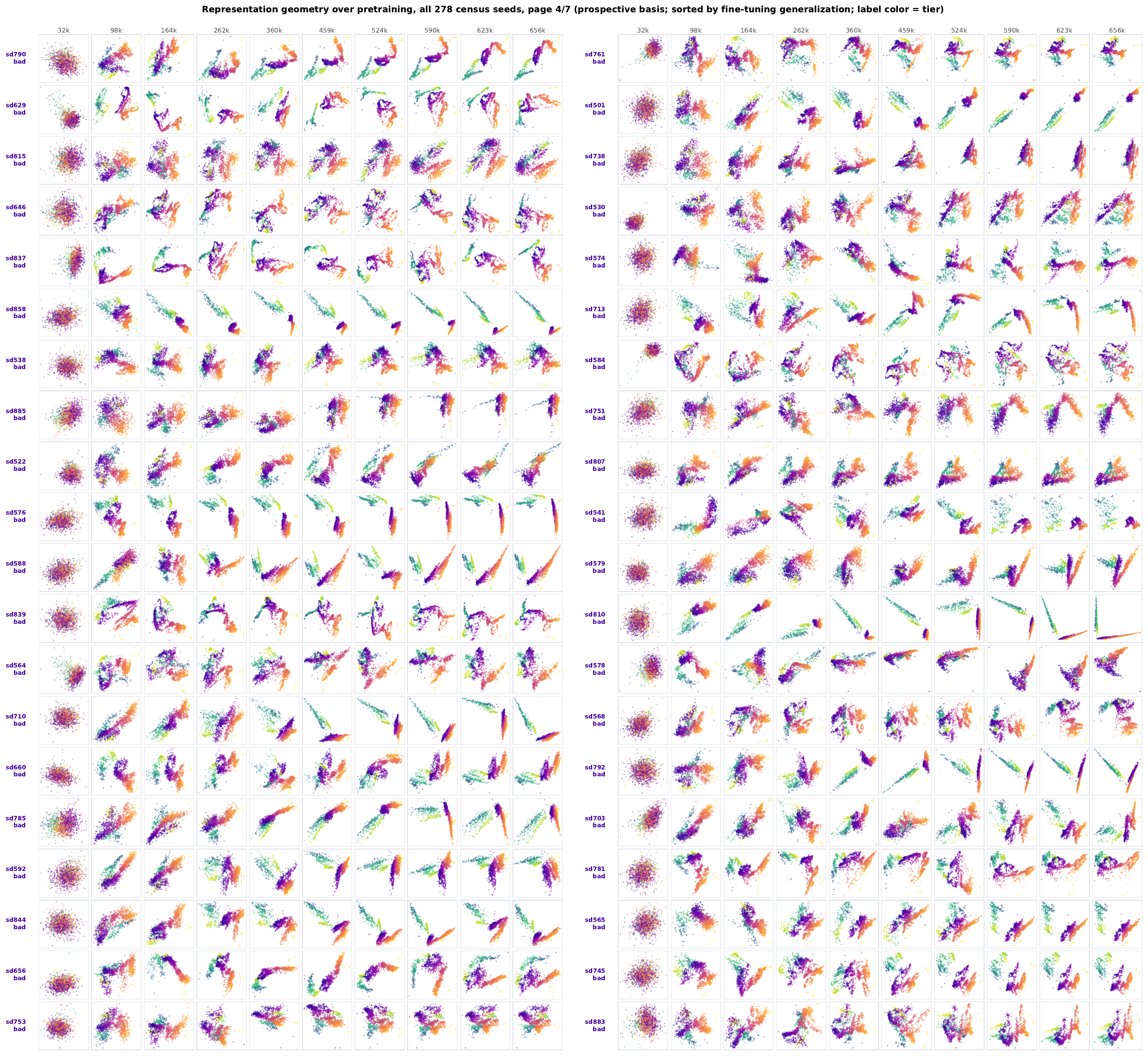}
\caption{\textbf{Representation maps across PT seeds (continued).} Additional seeds in the
layout of Fig.~\ref{fig:gallery-p1}.}
\label{fig:gallery-p4}
\end{figure}

\begin{figure}[p]
\centering
\includegraphics[width=\textwidth]{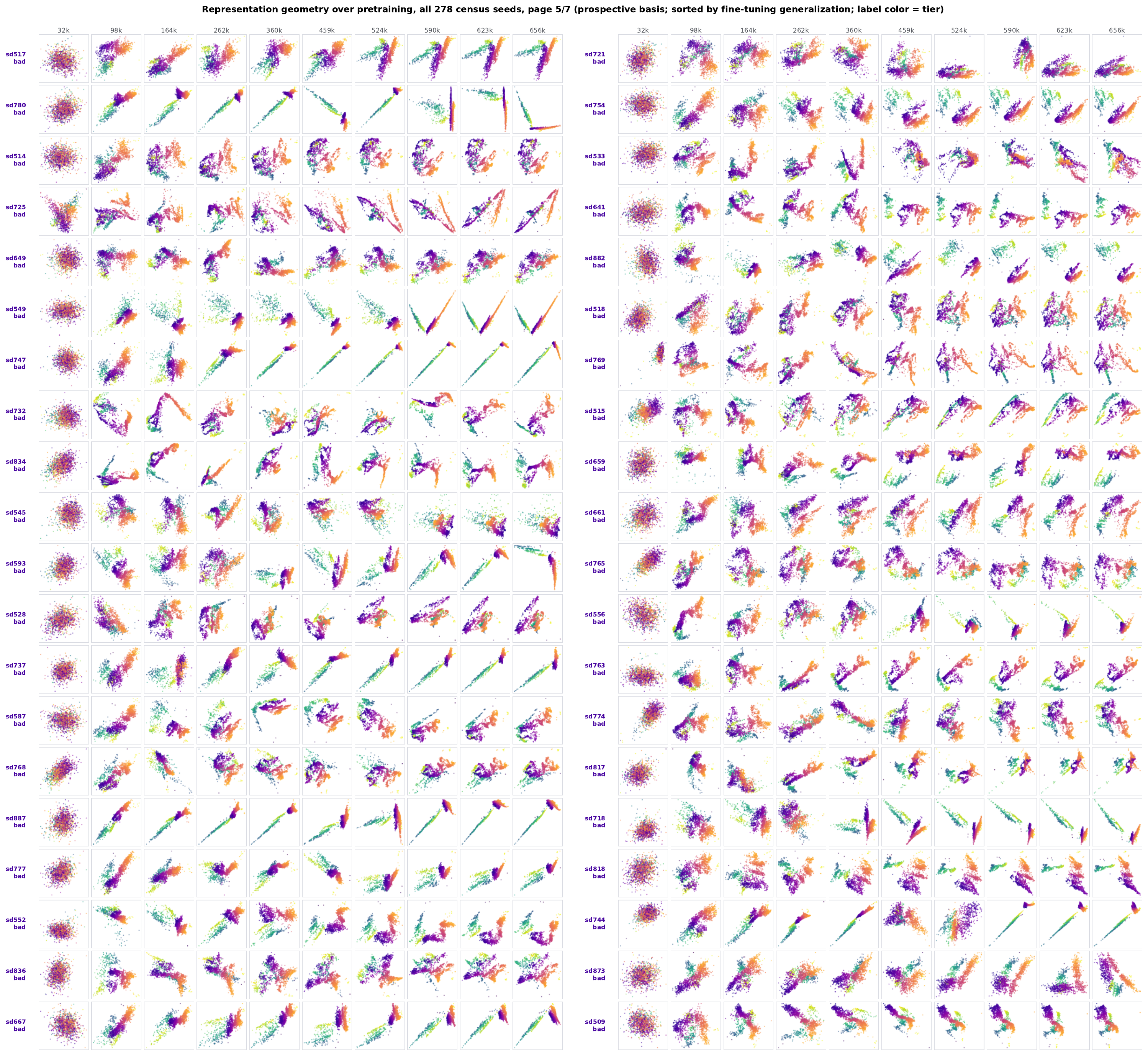}
\caption{\textbf{Representation maps across PT seeds (continued).} Additional seeds in the
layout of Fig.~\ref{fig:gallery-p1}.}
\label{fig:gallery-p5}
\end{figure}

\begin{figure}[p]
\centering
\includegraphics[width=\textwidth]{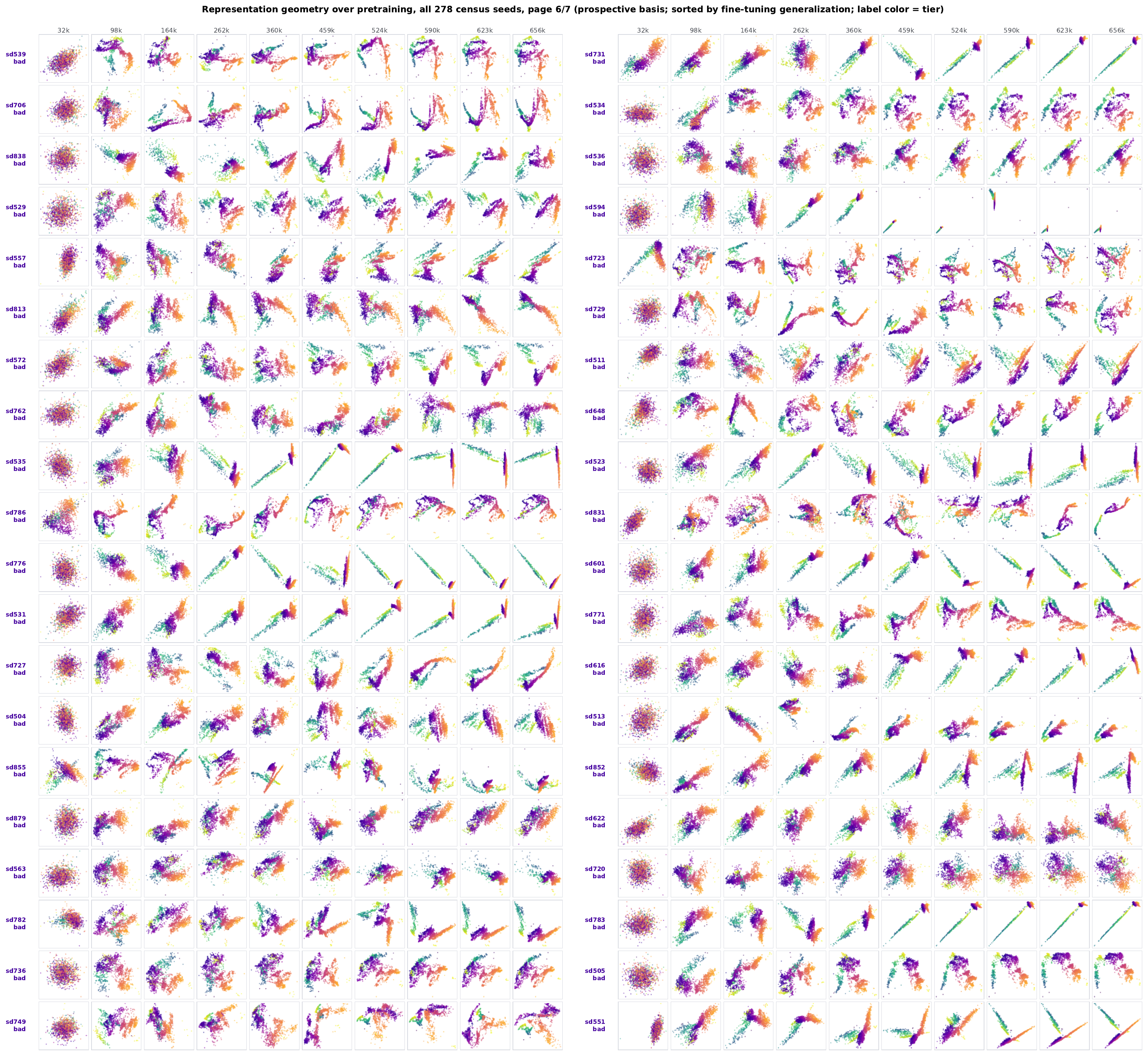}
\caption{\textbf{Representation maps across PT seeds (continued).} Additional seeds in the
layout of Fig.~\ref{fig:gallery-p1}.}
\label{fig:gallery-p6}
\end{figure}

\begin{figure}[p]
\centering
\includegraphics[width=\textwidth]{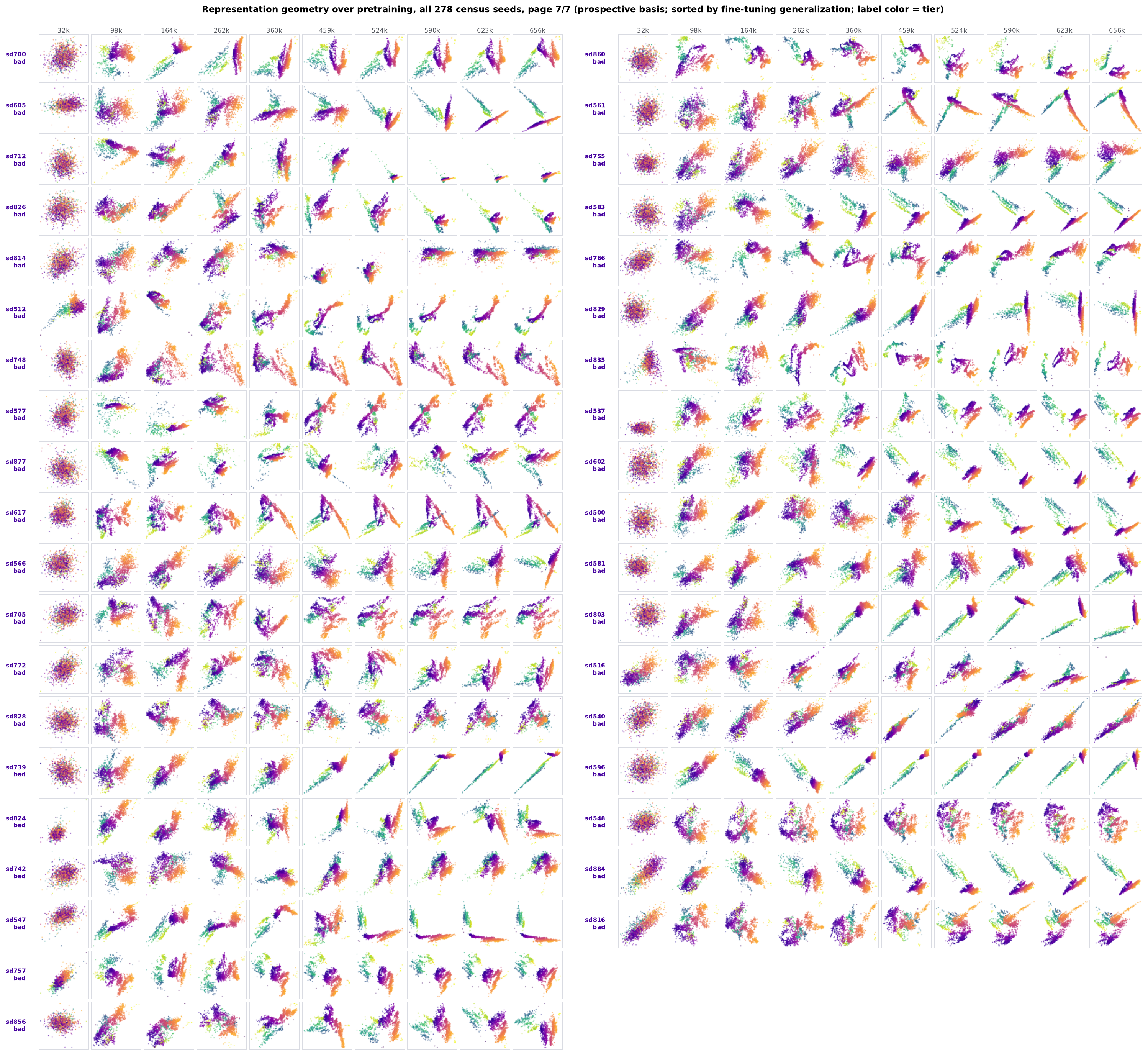}
\caption{\textbf{Representation maps across PT seeds (continued).} Additional seeds in the
layout of Fig.~\ref{fig:gallery-p1}.}
\label{fig:gallery-p7}
\end{figure}

Figs.~\ref{fig:gallery-p1}--\ref{fig:gallery-p7} display the layer-5 representations of
each seed across ten PT checkpoints, with each panel projected onto the basis fitted at that
checkpoint. Runs differing only in their random seed develop markedly different maps early
in PT, and these differences persist throughout training. This gallery provides a
qualitative visual illustration of the representational differences studied via
interventions in Sec.~\ref{sec:causal}.

\end{document}